# New directional bat algorithm for continuous optimization problems

Asma CHAKRI[1]*, Rabia KHELIF[2], Mohamed BENOUARET[3], Xin-She YANG[4],

*1. Industrial Mechanics Laboratory, Department of Mechanical Engineering, University Badji Mokhtar of Annaba (UBMA), BP12-23000, Annaba, Algeria*
*2. Industrial Mechanics Laboratory, Department of Mechanical Engineering, University Badji Mokhtar of Annaba (UBMA), BP12-23000, Annaba, Algeria*
*3. Industrial Mechanics Laboratory, Department of Mechanical Engineering, University Badji Mokhtar of Annaba (UBMA), BP12-23000, Annaba, Algeria*
*4. School of Science and Technology, Middlesex University London, The Burroughs, London NW4 4BT, United Kingdom*

**Abstract**

Bat algorithm (BA) is a recent optimization algorithm based on swarm intelligence and inspiration from the echolocation behavior of bats. One of the issues in the standard bat algorithm is the premature convergence that can occur due to the low exploration ability of the algorithm under some conditions. To overcome this deficiency, directional echolocation is introduced to the standard bat algorithm to enhance its exploration and exploitation capabilities. In addition to such directional echolocation, three other improvements have been embedded into the standard bat algorithm to enhance its performance. The new proposed approach, namely the directional Bat Algorithm (dBA), has been then tested using several standard and non-standard benchmarks from the CEC'2005 benchmark suite. The performance of dBA has been compared with ten other algorithms and BA variants using non-parametric statistical tests. The statistical test results show the superiority of the directional bat algorithm.

# 1 Introduction

Optimization in the real-world applications usually involves highly nonlinear complex problems with many design variables and complex constraints. The objective of an optimization problem can usually be associated with the minimization of wastes, costs and times, or maximization of benefits, profits and performance (Mazhoud, Hadj-Hamou, Bigeon, & Joyeux, 2013). Due to the fact that traditional deterministic methods or algorithms do not cope well to solve a large number of problems in practice, especially when the objective function is multimodal with many local optima, scientists have turned their eyes to the Mother Nature, looking for new ideas and inspiration for problem-solving. Since then, over a dozen algorithms have been developed based on the inspiration from different natural processes. Genetic algorithm (Davis., 1991) and differential evolution (DE)(Das & Suganthan, 2011) are based on the biological evolution processes. Cuckoo search (CS)(Yang & Deb, 2010) is based on the brooding behavior of some cuckoo species, while the firefly algorithm (FA) is based on the flashing patterns of tropical fireflies (Yang, 2010a). Particle swarm optimization (PSO) (Eberhart & Kennedy, 1995; Eberhart & Yuhui, 2001) and Ant colony optimization (ACO)(Dorigo, Maziezzo, & Colorni, 1996) are based on the swarm behavior. Harmony search (HS)(Geem, Kim, & Loganathan, 2001) is an algorithm inspired by the music composition process of musicians. These algorithms, often referred to as nature-inspired algorithms or bio-inspired algorithms, have become very popular due to their easy structure and their abilities to obtain a solution, while the traditional deterministic algorithms may fail (Zadeh & Shirazi, 2013). Most of these algorithms are swarm intelligence based algorithms because they try to mimic some of the key characteristics of swarming behavior of ants, birds, fish, insects and bats. Examples are PSO (Eberhart & Kennedy, 1995; Eberhart & Yuhui, 2001), ACO (Dorigo et al., 1996) and CS (Yang & Deb, 2009).

The bat algorithm (BA) proposed by Xin-She Yang (Yang, 2010b) is also a swarm intelligence based algorithm, inspired by the echolocation behavior of micro-bats. When flying and hunting, bats emit some short, ultrasonic pulses to the environment and list to their echoes. Studies show that the information from the echoes will enable bats to build a precise image of their surroundings and determine precisely the distance, shapes and prey's location. The capability of such echolocation of micro-bats is fascinating, as these bats can find their prey and discriminate different types of insects even in complete darkness (Yang, 2010b). The earlier studies showed that BA can solve constrained and unconstrained optimization problems with much more efficiency and robustness compared to GA and PSO (Gandomi, Yang, Alavi, & Talatahari, 2013; Yang, 2010b; Yang & Gandomi, 2012).

Although the bat algorithm was proposed recently, there are already several variants of BA in the literature (see the related work Section 3). Despite the fact that BA is a very powerful algorithm and can produce robust solutions on low dimensional problems, its performance diminishes significantly when the problem dimension increases, which is also true for other algorithms such as PSO and GA (X. Wang, W. Wang, & Y. Wang, 2013). Several studies reported that BA can be efficient to solve a diverse range of problems (Fister, Xin-She, Fong, & Yan, 2014); however, premature convergence can occur under certain conditions. Like all swarm intelligence based algorithms, the search mechanism of an algorithm is governed by two crucial components: exploitation and exploration. Briefly speaking, exploration is the ability of an algorithm to explore the search space and seeking for new, unknown regions, while exploitation is the ability to improve the solutions by exploiting information obtained from existing solutions.

The aim of this study is to improve the performance of the standard bat algorithm by increasing its exploration and exploitation abilities along the main line of the BA spirits. In this paper, new modifications and idealized rules have been embedded to the BA by using directional echolocation and other features. The proposed new directional bat algorithm (dBA) will be tested on several benchmark problems chosen from the well-known CEC'2005 benchmark set and compared with several other swarm and evolutionary algorithms. Therefore, this study is organized as follows. The standard bat algorithm is presented in Section 2, and a literature survey of different BA variants is presented in Section 3. Then, the new directional bat algorithm is described in Section 4. Finally, the results of the numerical experiments are presented in Section 5, followed by the discussions and conclusions in Section 6.

## 2 The standard bat algorithm

The standard Bat algorithm is inspired by the echolocation process of bats. By observing the behavior and characteristics of the micro-bats, Yang (Yang, 2010b) proposed the standard BA in accordance to three major characteristics of the echolocation process of the micro-bats. The used idealized rules in BA are:

a) All bats use echolocation to sense distance and the location of a bat $x_i$ is encoded as a solution to an optimization problem under consideration (Yang, 2010b).
b) Bats fly randomly with velocity $v_i$ at position $x_i$ with a varying frequency (from a minimum $f_{min}$ to a maximum frequency $f_{max}$) or a varying wavelength $\lambda$ and loudness $A$ to search for prey. They can automatically adjust the wavelengths (or frequencies) of their emitted pulses and the rate of pulse emission $r$ depending on the proximity of the target (Yang, 2010b).
c) Loudness varies from a large positive value $A_0$ to a minimum constant value $A_{min}$ (Yang, 2010b).

For each bat (*i*), its position (*x*$_i$) and velocity (*v*$_i$) in a *d*-dimensional search space should be defined. *x*$_i$ and *v*$_i$ should be subsequently updated during the iterations. The rules for updating the position and velocities of a virtual bat (*i*) are given as in (Yang, 2010b)

$$f_i = f_{min} + (f_{max} - f_{min})rand \tag{1}$$

$$v_i^{t+1} = v_i^t + \left(x^* - x_i^t\right)f_i \tag{2}$$

$$x_i^{t+1} = x_i^t + v_i^{t+1} \tag{3}$$

where *rand*∈[0,1] is a random vector drawn from a uniform distribution. Here $x^*$ is the current global best location (solution) which is located after comparing all solution among all the *n* bats. A new solution for each bat is generated locally using random walk given by Eq. (4)

$$x_{new} = x_{old} + \varepsilon <A^{t+1}> \tag{4}$$

where $\varepsilon \in [-1,1]$ is a random number, while $<A_i^{t+1}>$ is the average loudness of all the bats at this time step.

The loudness *A*$_i$ and the rate of pulses emission *r*$_i$ are updated as the iterations proceed. The loudness decreases and the pulse rate increases as the bat gets closer to its prey. The equation for updating the loudness and the pulse rate are:

$$A_i^{t+1} = \alpha A_i^t, \tag{5}$$

$$r_i^{t+1} = r_i^0 \left[1 - \exp(-\gamma t)\right], \tag{6}$$

where $0 < \alpha < 1$ and $\gamma > 0$ are constants. As $t \to \infty$, we have $A_i^t \to 0$ and $r_i^t \to r_i^0$. The initial loudness $A_0$ can typically be $A_0 \in [1, 2]$, while the initial emission rate $r^0 \in [0, 1]$.

The basic steps of the standard bat algorithm are summarized in the pseudocode as shown in Algorithm 1.

**3 Related works**

The standard bat algorithm has been proven to be a very powerful optimization tool and it can produce a robust solution on low-dimensional functions and a diverse range of applications (Alam & Kabir, 2014; Fister, Fister, & Yang, 2013; Mirjalili, Mirjalili, & Yang, 2013; Yilmaz & Küçüksille, 2015). Despite such progress, some studies reported that its performance may diminish as the problem's dimension increases (Fister et al., 2013). Such dimensionality-related issues are also true for almost all other algorithms, but there are some possible remedies that will be explored in this paper.

To overcome this deficiency, several methods and techniques were proposed. For example, chaotic maps can be used to enhance the performance and eleven chaotic map functions were used to find the best, potential chaotic-based bat algorithm (CBSO) variant (Jordehi, 2015). The best strategy found is to compute the loudness ($A_i$) by multiplying a linearly decreasing function by the iterative chaotic map with an infinite collapses function. From the results, the authors reported that CBSO could outperform the conventional BA, cuckoo search, Big bang-big crunch algorithm (BBBC)(Erol & Eksin, 2006), gravitational search algorithm (GSA)(Rashedi, Nezamabadi-pour, & Saryazdi, 2009), and genetic algorithm.

In addition, a binary bat algorithm (BBA) was also proposed (Mirjalili et al., 2013), and the BBA has an artificial navigating and hunting system in binary search spaces by changing their positions from "0" to "1" and vice versa, and their results showed the superiority of BBA to the binary PSO (BPSO) and genetic algorithm.

It was also shown that the addition of Gaussian perturbations and simulated annealing can also be useful (He, Ding, & Yang, 2014), and the authors introduced both simulated annealing (SA)(Kirkpatrick, Gelatt, & Vecchi, 1983) and Gaussian perturbation into the standard BA to speed up the global convergence rate (SAGBA). Once an initial population is generated, the best solutions are replaced by new solutions generated with SA equations, followed by the standard updating equations of BA. Then, the Gaussian perturbation process is used to perturb the locations/solutions and generate a set of new solutions. Results show that SAGBA performs better than the standard BA and hybrid method SA-PSO.

A novel bat algorithm based on differential operator and Lévy flights trajectory (DLBA) was also proposed (Xie, Zhou, & Chen, 2013), and the DLBA has the same structure of BA, but the movements of bats and local search are different. The differential operator which is similar to the mutation operation of differential evolution (DE) has been introduced to the frequency, making it fluctuate up and down and change self-adaptively. In addition, Lévy flight has also been incorporated to the local search equation to enable the bats jump out of any local optimum.

Based on the concepts of parallel processing and communication strategy, the authors in (Nguyen, Pan, Dao, Kuo, & Horng, 2014) presented a hybrid algorithm between the bat algorithm and artificial bee colony (BA-ABC). Each algorithm evolves independently, after running a fixed number of iterations, the two algorithms then exchange data. The bats with a solution near the best solution replace the worst artificial agents of ABC. In contrast, the better artificial agents of ABC are to replace the poorer bats of BA. In addition, the bats in BA do not know the existence of the artificial bees of ABC and vice versa. Results show that the hybrid BA-ABC increases the convergence and accuracy more than BA (up to 78%) and ABC (up to 11%).With the similar idea, Pan et al. (Pan, Dao, Nguyen, & Chu, 2015) used the parallel processing concept to hybridize particle swarm optimization with bat algorithm (PSO-BA).

Dao et al. (Dao, Pan, Nguyen, Chu, & Shieh, 2014) developed a compact version of the bat algorithm (CoBA), addressing to the hardware devices with limited resources such as the memory size or low price equipment. The bat population is replaced with a probability vector updated based on a single computation. These lead to an algorithm functioning with a modest memory usage. Results show that the CBSO performances are as good as the standard BA despite its modest memory usage.

The adaptive bat algorithm (ABA) was presented and each bat can adaptively adjust its flight speed and direction (X. Wang et al., 2013). They introduced an inertia factor to the velocity equation based on the distance between the actual bat position and the best bat position. They also proposed to add a shrinking factor to the local search formula to reduce the space search during the iterative process. With some similarity to ABA, the bat algorithm with recollection (RBA) was also proposed (W. Wang, Y. Wang, & X. Wang, 2013) and it has a time-varying velocity inertia weight factor which make the bats smoothly close to the current best position.

Chen et al. (Chen et al., 2014) introduced the Doppler Effect to the standard BA to improve its solving efficiency (DBA). The Doppler Effect produces a frequency shift caused by the velocity between the sound source and observer. They proposed to add the context-awareness concept to permit to the bat to sense the environmental change by physical laws, according to the Doppler Effect.

The hybrid bat algorithm (HBA) proposed by Fister et al. (Fister et al., 2013) has been adapted to become hybrid self-adaptive bat algorithm (HSABA) (Fister, Fong, Brest, & Fister, 2014). It has two main characteristics. First, they introduced a self-adaptation technique to the BA control parameters (loudness and pulse rate). Second, they replaced the local search equation by the differential evolution operator where four operator strategies were considered, namely *DE/rand/1/bin*, *DE/randToBest/1/bin*, *DE/best/2/bin* and finally *DE/best/1/bin*. Results showed that the hybrid bat algorithm based on "*DE/best/2/bin*" achieved the best result when compared with the other three strategies.

The enhanced bat algorithm (EnBA) proposed in (Yilmaz & Küçüksille, 2015) has been developed through three different methods. An inertia factor has been proposed to balance the search capabilities during the optimization process depending on the requirement of BA. A new equation to control the velocity of bats was introduced with an ability to contribute to the dispersion of the solutions into the search space. Finally, hybridization with invasive weed optimization (IWO)(Mehrabian & Lucas, 2006) was adopted to increase the exploitation capability rather than exploration with rectification toward the end of the optimization process.

The bat algorithm with self-adaptive mutation (BA-SAM) proposed in (Alam & Kabir, 2014) used Gaussian and Cauchy random mutation, instead of the original standard local search equation. The central idea for self-adaptation was to probabilistically select one of the two mutation schemes by using a learning strategy.

The modified bat algorithm (MBA) proposed by Yilmaz et al.(Yilmaz, Kucuksille, & Cengiz, 2014) has the same flowchart as the standard bat algorithm. A new strategies was proposed to update the control parameters (pulse rate and loudness), in addition to a new equation which was proposed to conduct a local search. Results showed a significant enhancement at low dimensional problems but not much in high dimensions.

Tsai et al. (Tsai, Pan, Liao, Tsai, & Istanda, 2011) proposed the evolved bat algorithm (EvBA) for solving numerical optimization problems based on the framework of BA. The EvBA was constructed with a new definition of the bats' movements. They introduced the sound speed to measure the distances and update the bats' movements.

Li and Zhou (Li & Zhou, 2014) used the complex-valued encoding to build the complex-valued bat algorithm (CVBA). The basic idea of the complex-valued encoding is to use two parameters, real part and imaginary part to represent a variable, and the real and imaginary parts can be updated in parallel. The independent variables of the objective function are determined by the modules and angles of their corresponding complex number. Therefore, the diversity of population is greatly enriched.

All of these algorithms have been tested on the classical benchmark functions. The majority of these function have the optimum located in the centre of the space search with null vector ($x=[0...0]^d$) as a solution, like the Griewank function, Rastrigin function, Ackley function, etc. Some of them showed a significant enhancement over the standard bat algorithm, but none of them had been tested against non-standard benchmark functions, like the CEC'2005 benchmark suite (Suganthan et al., 2005) where the functions have been shifted, rotated and displaced in order to ensure that their optima can never be found in the centre of the search space.

This paper attempts to improve the bat algorithm from a different perspective by mainly using the directional feature of echolocation. First, movement of bats should be directed by other and better bats, and the local movements can be refined by controlling the step sizes. Second, the rates of pulse emission and loudness are also modified to potentially enhance the performance. More specifically, four different modifications will be introduced to improve the efficiency of the bat algorithm.

**4 The new directional bat algorithm**

The new directional bat algorithm has the same procedure or flowchart as the standard bat algorithm. But we will introduce four modifications with the aim to enhance the exploitation and exploration capabilities of the bat algorithm so as to improve the BA performance.

4.1 The 1$^{st}$ modification (the directional echolocation)

The directional echolocation is used by bats as a main navigation system. During their flights, bats emit continuously short pulses that lasts a few milliseconds. By analyzing the echoes, they can create a 3D mental image of their surroundings. The information of other bats such as their positions can be useful to guide the search process. In addition, pulses can travel in different directions, and thus it may be useful to assume that each bat emits two pulses into two different directions before deciding in which direction that it will fly and this feature can be used to simulate the time difference or delay between echoes received by two ears. To extend along this line of thinking further, it can also be assumed that all the bats emit a pulse in the direction of the best bat (solution) where the food is considered to exist, and the other pulse to the direction of a randomly chosen bat.

As shown in Fig. 1, a bat emits two pulses in two different directions, one in the direction of the bat with the best position (the best solution), and the other to the direction of randomly selected bat. From the echoes (feedback), the bat can know if the food exists around these two bats or not. The best position is determined by the objective fitness, while, around the randomly selected bat, it depends on its fitness value. If it has a better fitness value as the actual bat, then the food is considered to exist, otherwise there is not a food source in the neighborhood.

If the food is confirmed to exist around the two bats (case 1), the current bat moves to a direction at the surrounding neighborhood of the two bats where the food is supposed to be plenty. If not (case 2), it moves toward the best bat. The mathematical formulas of the bats' movements are thus:

$$\begin{cases} x_i^{t+1} = x_i^t + (x^* - x_i^t)f_1 + (x_k^t - x_i^t)f_2 & (if\ F(x_k^t) < F(x_i^t)) \\ x_i^{t+1} = x_i^t + (x^* - x_i^t)f_1 & Otherwise \end{cases} \qquad (7)$$

where $x_k^t$ is the location of randomly selected bat ($k \neq i$) and $x^*$ is the best solution. The $F(.)$ is fitness function, while $f_1$ and $f_2$ are the frequencies of the two pulses and updated as follows:

$$\begin{cases} f_1 = f_{min} + (f_{max} - f_{min})rand1 \\ f_2 = f_{min} + (f_{max} - f_{min})rand2 \end{cases} \qquad (8)$$

Both $rand1$ and $rand2$ are two random vectors drawn from a uniform distribution between 0 and 1.

The directions of the movement generated by Eqs. (1-3) are directed towards the bat with the best position. This mechanism allows the BA to exploit more around the best position; however, if the best bat is not near the global optimality, there is a risk that the solutions generated by such moves could be trapped in local optima if the moves are not far enough. In this case, the algorithm can hardly escape local optima, which can lead to a premature convergence. The proposed movement in Eq. (7) has the ability to diversify the movement directions which can enhance the exploration capability, especially at the initial stages of iterations, and can thus avoid premature convergence. Furthermore, when it approaches the end of the iteration process, bats tend to gather around the best bat with stronger exploitation ability, which can in turn reduce the distances between them and consequently enhance the speed of convergence. Thus, Eq. (7) can promote different capabilities at different stages of iterations, leading to a better search mechanism and improved performance.

4.1 The 2$^{nd}$ modification

The second modification concerns the local search mechanisms. In the standard BA, bats are allowed to move from their current positions to new random positions using a local random walk. Here, we modify this move by the following equation:

$$x_i^{t+1} = x_i^t + <A^t> \varepsilon w_i^t \qquad (9)$$

where $<A^t>$ is the average loudness of all bats and $\varepsilon \in [-1,1]$ is a random vector. Here, $w_i$ is a parameter that can regulate the scales of the search as the iterative process proceeds. It starts from a large value (about a quarter of the typical scaling of the search domain) and then decreases to around 1% of the quarter of this length. We found that the following monotonically decreasing function is more suitable and gives stability to the algorithm:

$$w_i^t = \left( \frac{w_{i0} - w_{i\infty}}{1 - t_{\max}} \right)(t - t_{\max}) + w_{i\infty} \qquad (10)$$

Here, $w_{i0}$ and $w_{i\infty}$ are the initial and final values, respectively. In essence, $w_i$ can control the iteration procedure. In general, we can set $w_{i0}$ and $w_{i\infty}$ as follows:

$$w_{i0} = (Ub_i - Lb_i)/4 \qquad (11)$$

$$w_{i\infty} = w_{i0}/100 \qquad (12)$$

where $t$ is the current iteration and $t_{\max}$ is the maximum number of iterations. Here, $Ub_i$ and $Lb_i$ are the upper and lower bounds, respectively.

At the beginning of the iterative process, $w_i$ starts with a large value. It allows the bats to move randomly so as to increase the exploration ability of the algorithm and thus be able to explore the whole search space more effectively. At the end of the iterative process, the value of $w_i$ decreases, which reduces the search region around the best solution, and thus the exploitation capability of the algorithm is also enhanced.

In addition, from a pre-experiment analysis, we found that the assignment of a larger value to $w_{i0}$ may slow down the convergence of the algorithm. This is due to the fact that a large number of the generated solutions by Eq. (9) can be too far away and even outside of the domain. By using the sample technique such as Monte Carlo simulation with $10^6$ sampling, we found that in case of $w_{i0} = (Ub_i - Lb_i)$, at the first iteration, about 50% of the solutions generated by Eq. (9) are outside of the domain bounds. In case where $w_{i0} = (Ub_i - Lb_i)/2$, approximately 25% of the solution are outside of the bounds, while in case of $w_{i0} = (Ub_i - Lb_i)/4$, only about 12.5% are out of bounds. Assigning a low value to $w_{i0}$ can reduce the number of solutions out of bounds (for $w_{i0} = (Ub_i - Lb_i)/10$, about 5% of solutions are outside of the domain). However, it reduces also the probability to discover an optimal solution at the nearby of the boundaries (which is the case for certain problems). Therefore, we have used the values as given in Eqs.(11) and (12), based on the parametric studies using sampling techniques.

4.3 The 3$^{rd}$ modification

Equations (5) and (6) proposed by Yang (Yang, 2010b) to update the pulse rate and loudness to reach their final value during the iterative process very quickly, thus reducing the possibility of the auto-switch from global search to local search due to a higher pulse rate, and the acceptance of a new solution (low loudness). Therefore, we propose here to use these monotonically increasing, decreasing, pulse rate and loudness in the following form:

$$r^t = \left(\frac{r_0 - r_\infty}{1 - t_{max}}\right)(t - t_{max}) + r_\infty \qquad (13)$$

$$A^t = \left(\frac{A_0 - A_\infty}{1 - t_{max}}\right)(t - t_{max}) + A_\infty \qquad (14)$$

where the index 0 and ∞ stand for the initial and final values, respectively.

Loosely speaking, the pulse rate controls the movements of the bats by switching between Eq. (9) (local search) and Eq. (7) (global search). At the beginning of iterations, dBA tends to promote global search over local random walks so as to explore the search space more effectively. This mechanism is obtained by attributing a low value to $r_0$. However, this value should not be too low, thus allowing to a small fraction of bats to exploit the solutions of the bat with the good positions. While the iterations approach the end, a large value should be assigned to the pulse rate so that exploitation takes over from exploration. The loudness $A$ controls the acceptance or rejection of a new generated solution. The importance of this parameter is that by rejecting some solutions, it allows the algorithm to avoid being trapped in local optima (and thus avoid the premature convergence as well). Therefore, based on some pre-experiment studies, we recommend the following settings of the pulse rate and loudness: $r_0 = 0.1$, $r_\infty = 0.7$, $A_0 = 0.9$ and $A_\infty = 0.6$.

4.4 The 4$^{th}$ modification

The final improvement we made to the original BA is to allow the bats to update the pulse rate and loudness, and to accept a new solution if their movement produces a solution better than the old one, instead of the global best solution as in the standard BA. This modification was also suggested by other researchers (Hasançebi, Teke, & Pekcan, 2013). In addition, the acceptance of a new solution requires the fulfilling of two conditions. First, the solution has to produce an objective value lower than the current one (for the minimization problems). Second, a randomly generated number has to be lower than the current corresponding loudness. There exists a probability that the movement of the bat produces a solution better even to the global best solution and cannot be accepted because the randomly generated number is higher than the current loudness, especially at the end of the iterative process where the value of loudness is lower. Therefore, we can here allow the algorithm to update the global best position whenever the bat's random walk produces a solution with a better fitness value even if it was not accepted to update the bat's position. To summarize the above modifications, the pseudo-code of the new bat algorithm is illustrated in Algorithm 2.

## 5 Experiments and discussions

To validate the performance of the proposed directional bat algorithm, we have carried out various numerical experiments, which can be summarized as three comparison experiments. The first one is a comparison between the new directional bat algorithm and the standard algorithms including the bat algorithm on the classical benchmark functions. For the second one, the CEC'2005 benchmark suite (Suganthan et al., 2005) has been considered, and a comparison has been performed against some advanced optimization algorithms such as Self-adaptive Differential Evolution (SaDE)(Qin & Suganthan, 2005). The last comparison consists of a comparison with some improved bat algorithms on classical benchmark functions.

5.1 Benchmarking and Parameter Settings

Twenty popular benchmark functions shown in Table 1 have been used to verify the performance of the new bat algorithm, compared with that of standard BA, PSO, HS, CS, GA, and DE. The description and the setting parameters of these algorithms are as follows:

- **dBA:** An extensive analysis was performed to carry out parameter settings of dBA, for best practice, we recommend the following settings: $r_0 = 0.1$, $r_\infty = 0.7$, $A_0 = 0.9$, $A_\infty = 0.6$, $f_{min} = 0$ and $f_{max} = 2$.
- **BA:** The standard bat algorithm was implemented as it is described in (Yang, 2010b) with $r_0 = 0.1$, $A_0 = 0.9$, $\alpha = \gamma = 0.9$, $f_{min} = 0$ and $f_{max} = 2$.
- **PSO:** A classical particle swarm optimization (Eberhart & Kennedy, 1995; Eberhart & Yuhui, 2001) model has been considered. The parameters setting are $c_1 = 1.5$, $c_2 = 1.2$ and the inertia coefficient $w$ is a monotonically decreasing function from 0.9 to 0.4.
- **HS:** The considered harmony search algorithm is the standard one described in (Geem et al., 2001) with the following setting $BW = 0.2$, $HMCR = 0.95$, $PAR = 0.3$.
- **CS:** The cuckoo search via Lèvy flights describe in (Yang & Deb, 2009) is considered with the probability of the alien eggs discovery $p_a = 0.25$.

- **GA:** standard genetic algorithm (Davis., 1991) with a crossover probability = 0.95 and mutation probability = 0.05.
- **DE:** The classical differential evolution as described in (Das & Suganthan, 2011) with "DE/rand/1/bin" strategy is considered. The parameters setting are $CR = rand[0.2, 0.9]$ and $F = rand[0.4, 1]$.

For a fair comparison, the common parameters are considered the same. The population size was set to $N = 30$, and the number of function evaluations is the same as 15000, without counting the initial evaluations, though all algorithms were initialized randomly in the similar manner. Therefore, we set $t_{max} = 500$ except for CS. Due to the fact that the CS algorithm uses a number of $2N$ function evaluations at each iteration, we adjust $t_{max}$ for this case to 250. The dimensionality of the all benchmark function is $D = 30$.

5.2 The first Experiment

For meaningful statistical analysis, each algorithm was run 51 times using a different initial population at each turn. The global minimum obtained after each trial was recorded for further statistical analysis. Subsequently, the mean-value of the global minimum, the standard deviation (SD), the best solution, the median and the worst solution values have been computed and presented in Tables 2 and 3.

From the results presented in Tables 2 and 3, the new directional bat algorithm achieved better results for 9 functions (F1, F3, F4, F10, F12, F14, F15, F18 and F19) and it obtained the best median for F2 and best solution for F16. While the GA obtained better results for 5 functions (F6, F13, F16, F17, and F20). The DE has better scores for 3 functions (F5, F7 and F8). HS obtained best results for F9 and BA for F11.

Figures 2 and 3 represent the evolution of the mean of the minimum obtained at each iteration for 51 trials of the classical benchmark functions. At the beginning of the iterative process, and due to the low value of the pulse rate, the algorithm promotes the exploration phase which enables to the dBA to explore a large area of the search space. While the iterative process continues, the pulse rate value increases, which maintains a balance between exploitation and exploration. Near the end of the iterations, the high value of the pulse rate fosters the exploitation phase which enables the algorithm to enhance the quality of the obtained results and accelerate the convergence rate. The comparison with the other algorithms shows the superiority of dBA in several benchmarks.

To evaluate dBA's performance, non-parametric statistical tests were carried out. We performed pairwise and multiple comparisons ($1 \times N$). The considered pairwise comparisons are: the Sign test and Wilcoxon's (Derrac, García, Molina, & Herrera, 2011). For the multiple comparisons we use the Friedman's test, the Aligned Friedman's test and the Quade test with the following associated post-hoc procedures: Holland, Rom, Finner and Li (Derrac et al., 2011; García, Molina, Lozano, & Herrera, 2009).

Table 4 presents the pairwise comparison results. The first row presents the number of wins and losses of dBA versus the rest of the algorithms on each problem listed in Table 1. The algorithm is considered a winner if the mean of 51 runs on a single problem is better than the other algorithm. Since we have 20 problems, an algorithm is consider better with a level of significance $\alpha = 0.05$ if it has 15 wins at least (see Table 4 in (Derrac et al., 2011)). From the results, dBA significantly outperforms the standard BA, PSO, HS, and CS. The second row represents the *p*-value of the Sign test. As it can be seen, dBA is superior to BA, PSO, HS, and CS. The Wilcoxon *p*-values presented in the last row shows that dBA outperforms BA, PSO, HS, CS and GA with level of significance $\alpha = 0.05$ and DE with $\alpha = 0.1$.

For the multiple comparisons, the results are obtained using CONTROLTEST java package proposed by Derrac et al. (Derrac et al., 2011) and downloaded from the SCI2S website (www.sci2s.ugr.es/sicidm/). Table 5 presents the Friedman, Aligned Friedman and the Quade ranks. For each test, an algorithm is considered better if it has a low rank. From the results, dBA has the lowest rank for the three tests, which means that it is the best performing algorithm from the comparison. In addition, the last two rows present the statistic and *p*-value of each test. For Friedman and Aligned Friedman, the statistic is distributed according to the chi-square distribution with 6 degrees of freedom, while for the Quade test, the static is distributed according to the *F*-distribution with 6 and 114 degrees of freedom. The low *p*-value of the different tests suggest the existence of significant differences among the considered algorithms (Derrac et al., 2011).

To highlight the differences between algorithms, Table 6 presents the *z*-values, the unadjusted *p*-values and the adjusted *p*-values with different post-hoc procedures of the Friedman, Aligned Friedman and Quade tests. The control method is dBA. The *z*-value in all cases is used to find the corresponding *p*-value on the normal distribution N(0,1) (Derrac et al., 2011). The analysis of the unadjusted p-value of the Friedman test shows significant differences between dBA and five algorithms (PSO, BA, HS, CS and GA), while the Aligned Friedman test considers the existence of differences with PSO, HS and BA only. The Quade test reveals that dBA is significantly superior to BA, HS, PSO and GA.

In (Derrac et al., 2011), the authors reported that the unadjusted *p*-values are not suitable for multiple comparisons due to the family error accumulation; therefore, they suggested to use the adjusted *p*-value. Several post-hoc procedures were proposed, we selected the most powerful method among them. The adjusted p-values of the Friedman test show that dBA outperforms significantly PSO, BA, HS, CS and GA, while the aligned Friedman *p*-values outlines the differences only with BA, HS and PSO. Knowing that the Quade test takes into account the relative difficulties of problems, the corresponding adjusted *p*-values suggest that dBA achieves better results in harder problems than HS, PSO and BA, and behave similarly or better for most difficult problems to CS, GA and DE.

The so-called contrast estimation is used to measure the differences between the performance of algorithms. It assumes that the expected differences between algorithms' performance are the same across problems; thus, the capability of an algorithm is reflected by the magnitudes of the differences between them (Derrac et al., 2011). This procedure is considered a safer metric in multiple-comparison between algorithms (García, Fernández, Luengo, & Herrera, 2010), and can be used to estimate by how far an algorithm outperforms another one (Derrac et al., 2011). The contrast estimation results are summarized in Table 7. Each row presents the comparison results between the corresponding algorithm in the first column and the rest of the algorithms. If the estimated value is positive, it means that the algorithm in first column is better and vice versa if the value is negative. In addition, a high estimated value means there are high differences between the algorithms. By analyzing the row for dBA, we can see that it always obtains a positive difference value, which means that it outperforms the rest of all the algorithms.

5.3 The 2$^{nd}$ experiment

In this second experiment, the CEC'2005 benchmark suite is tested. It is composed of 25 different functions with different characteristics such as scalability, separability and multimodality. All the functions of the CEC'2005 suite are generated from the basic functions listed in (Suganthan et al., 2005) by shifting, rotating or hybridizing these functions. These operations add more complexity to the problems. Descriptions of these functions are listed in Table 8. These benchmark functions are available at www.ntu.edu.sg/home/EPNSugan.

Following the common criteria presented in (Suganthan et al., 2005), all the benchmark functions have been executed 25 times each. The dimension is set to $D$=10 and the termination criteria is 100000 of function evaluations. The function error ($f(x) - f(x^*)$) value is recorded after each run and ranked from the smallest to the largest. The function errors are the differences between the best results obtained by dBA and the true global optima (Table 8). Table 9 presents the 1$^{st}$ (best), the 7$^{th}$, the 13$^{th}$ (median), the 19$^{th}$ and the 25$^{th}$ (worst) function error values, in addition to the mean and the standard deviation of 25 runs for all the special benchmark functions of the CEC'2005 suite.

To evaluate the performance of dBA in comparison with others algorithm, Derrac et al. (Derrac et al., 2011) presented a practical tutorial on the use of non-parametric statistical tests for comparing swarm and evolutionary algorithms. They used the CEC'2005 benchmark suite to illustrate the use of the tests for pairwise and multiple comparisons. A set of well-known evolutionary and swarm intelligence algorithms have been used for comparison. We will use the results of these algorithms provided by (Derrac et al., 2011) to carry out the comparison of the dBA performance and the other algorithms on CEC'2005 benchmark problems. The considered algorithms are: the PSO algorithm (Derrac et al., 2011), restart covariant matrix evolutionary strategy with increasing population (IPOP-CMA-ES) (Auger & Hansen, 2005; Derrac et al., 2011), the CHC algorithm (Derrac et al., 2011; Eshelman, 1991; Eshelman & Schaffer, 1993), steady-state genetic algorithm (SSGA) (Derrac et al., 2011; Fernandes & Rosa, 2001;

Mühlenbein & Schlierkamp-Voosen, 1993), two instances of the classic scatter search model SS-Arit and SS-BLX (Derrac et al., 2011; Herrera, Lozano, & Molina, 2006; Laguna & Marti, 2002), the classical differential evolution with two crossover strategies *Rand/1/exp* (DE-Exp) and *Rand/1/bin* (DE-Bin) (Derrac et al., 2011; Price, Storn, & Lampinen, 2005) and finally Self-adaptive differential evolution (SaDE) (Derrac et al., 2011; Qin & Suganthan, 2005). The IPOP-CMA-ES is the winner of the CEC'2005 competition on real parameter optimization. The parameter settings of the above algorithms are described in (Derrac et al., 2011).

In earlier studies by (García et al., 2009), they reported that running an algorithm for 25 times for a single problem may be low for carrying out statistical analysis, but it was a requirement in the CEC'2005 Special Session (Suganthan et al., 2005). Therefore, authors in (Derrac et al., 2011) have run each algorithm listed in the previous paragraph 50 times for a significant analysis. Consequently, we have executed the dBA 50 times on each single problem listed in Table 7. The parameter settings of dBA are the same as mentioned in Section 5.1.

Table 10 presents a comparison of the average errors obtained for the 25 benchmark functions with dBA and the other algorithms. The algorithms are compared based on the mean values for 1E+5 function evaluations with dimension $D = 10$. Results are within three digits of precision. If the error is less than 1E-10, it is considered as 0E+00. As it can be seen, dBA has outperformed or performed equally best as the other algorithms for twelve functions: SF01, SF02, SF05, SF07, SF13, SF18 and SF20-25 which represents 48% of the problems.

The pairwise comparisons results are presented in Table 11. It presents the number of wins, ties and losses, the Sign test *p*-values, the level of significance according to the number of wins ($\alpha$) and the Wilcoxon *p*-values. The analysis of the Sign test *p*-values reveal that dBA is significantly better than PSO, CHC, SSGA, SS-BLX, SS-Arit and the winner of the CEC'2015 congress IPOP-CMA-ES. According to the number of wins and the number of problems (25 benchmark functions), an algorithm is considered better than another with a level of significance $\alpha = 0.05$ if it has 18 wins and with $\alpha = 0.1$ if it has 17 wins. We note that the number of ties is divided between the two algorithms equally, and if there is an odd number of them, one should be ignored. Therefore, dBA outperforms PSO, IPOP-CMA-ES, CHC, SSGA and SS-Arit with $\alpha = 0.05$, and also DE-Bin, DE-Exp and SaDE with $\alpha = 0.1$. The Wilcoxon *p*-values show the superiority of dBA over the PSO, IPOP-CMA-ES, CHC, SSGA, SS-BLX and SS-Arit.

Table 12 presents the Friedman, Aligned Friedman and Quade ranks of the ten algorithms. As it can be seen, dBA has the lowest rank of the three tests. The statistic of the Friedman and the Aligned Friedman test is distributed according to chi-square with 9 degrees of freedom, and the statistic of the Quade test is according to the F-distribution with 9 and 216 degrees of freedom. The low *p*-values of the three tests suggest the existence of significant differences between the algorithms.

The Adjusted *p*-values of the three tests are presented in Table 13. The analysis of the Friedman adjusted *p*-values shows the superiority of dBA over PSO, CHC, SSGA, SS-Arit, SS-BLX and IPOP-CMA-ES according to Finner and Li post-hoc procedures, while the Holland and Rom procedures exclude the IPOP-CMA-ES. The adjusted *p*-values of Aligned Friedman tests imply the superiority of dBA over CHC, SSGA and PSO with $\alpha = 0.05$. The Li post-hoc procedures include also the IPOP-CMA-ES algorithm. The Quade test *p*-values highlight only the differences between dBA and CHC. From the results, we can conclude that dBA behaves in the same way as or better than SaDE, DE-Bin and DE-Exp.

The contrast estimation results are presented in Table 14. As it can be seen in the first row, the directional bat algorithm obtained positive difference values with respect to the other algorithms. This indicates that dBA outperforms the remaining algorithms. The rest of the results pinpoint SaDE as the second best algorithm, while CHC algorithm has the worst performance among them.

5.4 The 3$^{rd}$ experiment

The third experiment consists of comparisons of the dBA performance with 6 variants of the bat algorithm that exist on the literature, to be precise the CBSO(Jordehi, 2015), BBA(Mirjalili et al., 2013), SAGBA(He et al., 2014), HSABA(Fister, Fong, et al., 2014), EnBA(Yilmaz & Küçüksille, 2015) and MBA(Yilmaz et al., 2014). Six classical benchmark functions, which were commonly used in different BA variants, are considered, namely the spherical function, Griewank function, Rastrigin's function, Ackley's function, Rosenbrock's function and Zakharov function.

Each function was evaluated in different situations of bounds, dimensions, populations, maximum number of iterations and number of trials. Results are presented in Table 15. As it can be seen, dBA outperforms the other algorithms in 26 situations from 32, which represents 81.25% of benchmarking scenarios.

The most recently proposed algorithm CBSO(Jordehi, 2015) has been outperformed by dBA in 3 situations of 4. To overcome the exploration deficiency of CBSO, the author used a large number of bat individuals which increases the number of function evaluations. Therefore, for the same number of function evaluation which is 1E+6, we have executed the dBA with different settings of population sizes and numbers of maximum iterations. For each situation, we have run dBA 30 times, and we have computed the mean, the standard deviation and the success rate. Since the global optima of the four functions are all 0 (spherical, Griewank, Rosenbrock and Zakharov's function), we can consider a run successful if the outcome is less the 1E-10. Results are presented in Table 16.

From the analysis of the min/max solutions of 30 runs obtained by CBSO, none of the runs has been successful. On the other hand, dBA has 100% of success in the minimization of the spherical function for different values of bat populations and numbers of maximum iterations.

For Zakharov's function, dBA obtains 100% of success rate for a population of bats less or equal to 100 individuals. The success rate of the minimization of Griewank's function is between 30% and 37%. For Rosenbrock's function, no run was successful and the best solution is obtained by CBSO.

Through the extensive experiments and benchmarking, we can conclude that dBA is superior to other algorithms in terms of accuracy and search efficiencies. The four modifications introduced in the proposed approach can indeed enhance and improve the performance of the bat algorithm. This provides a basis for further studies in solving real-world applications.

**6 Conclusions**

In this study, an improved version of the standard bat algorithm, called the new directional bat algorithm, has been proposed and presented. Four modifications have been embedded to the BA to increase its exploitation and exploration capabilities and consequently have significantly enhanced the BA performance. Three sets of experiments have been carried out to prove the superiority of the proposed dBA. In the first experiment, 20 classical benchmark functions have been used. Results have been compared with those obtained by the standard algorithms, namely PSO, HS, CS, GA, DE in addition to BA. The non-parametric statistical comparisons prove the superiority of dBA. In the second experiment, the CEC'2005 benchmarks suite was considered. dBA's results have been compared with other advanced swarm and evolutionary algorithms, the nonparametric statistical tests have showed that dBA had outperformed some of the stat-of-the-art algorithms such as the SaDE and IPOP-CMA-ES. The third experiment reveals that dBA outperforms many variants of BA that exist in the literature.

The main modifications and enhancements about the bat algorithm are to use two sets of pulse emissions in two different directions, which have led to a more efficient algorithm. The use of directions of best bats as higher-level information to guide new search becomes beneficial to enhance both the exploration and exploitation capabilities because it directs the moves more onto the promising regions. In addition, the variations and control of the pulse emission rates and the loudness provides an adaptive mechanism to control exploration and exploitation at different stages of iterations. At the earlier stage of iterations, search moves are mainly explorative, while search becomes more extensive and local at later iterations. It can be expected that further modifications with multiple pulse emissions along multiple directions can be worth investigating and this can form a useful topic for further research. In addition, to test the proposed approach against even higher-dimensional problems can also be very useful. Furthermore, it can be fruitful to apply the new directional bat algorithm to solve challenging problems in real-world applications.

**Algorithm 1**
The standard bat algorithm.

| | |
|---|---|
| 1. | Define the objective function |
| 2. | Initialize the bat population $-Lb_i \leq x_i \leq Ub_i$ ($i = 1,2,..,n$) and $v_i$ |
| 3. | Define frequencies $f_i$ at $x_i$ |
| 4. | Initialize pulse rates $r_i$ and loudness $A_i$ |
| 5. | *While* ($t \leq t_{max}$) |
| 6. |     Adjust frequency Eq. (1) |
| 7. |     Update velocities Eq. (2) |
| 8. |     Update locations/solutions Eq. (3) |
| 9. |     *if* (*rand* > $r_i$) |
| 10. |         Select a solution among the best solutions |
| 11. |         Generate a local solution around the selected best solution Eq. (4) |
| 12. |     *end if* |
| 13. |     Generate a new solution by flying randomly |
| 14. |     *if* (*rand* < $A_i$ & $F(x_i) < F(x^*)$) |
| 15. |         Accept the new solutions |
| 16. |         Increase $r_i$ Eq.(5) |
| 17. |         Reduce $A_i$ Eq. (6) |
| 18. |     *end if* |
| 19. |     Rank the bats and find the current best $x^*$ |
| 20. | *end while* |
| 21. | Output results for post-processing |

**Algorithm 2**
The new directional bat algorithm.
1.     Define the objective function
2.     Initialize the bat population $Lb_i \leq x_i \leq Ub_i$ ($i=1,2,..,n$)
3.     Evaluate fitness $F_i(x_i)$
4.     Initialize pulse rates $r_i$ loudness $A_i$ and $w_i$
5.     *While* ($t \leq t_{max}$)
6.         Select a random bat ($k \neq i$)
7.         Generate frequencies Eq. (8)
8.         Update locations/solutions Eq. (7)
9.         if ($rand > r_i$)
10.             Generate a local solution around the selected solution Eq. (9)
11.             Update $w_i$ Eq. (10)
12.         end if
13.         if ($rand < A_i$ & $F(x_i^{t+1}) < F(x_i^t)$)
14.             Accept the new solutions
15.             Increase $r_i$ Eq. (13)
16.             Reduce $A_i$ Eq. (14)
17.         end if
18.         if ($F(x_i^{t+1}) < F(x^*)$)
19.             Update the best solution $x^*$
20.         end
21.     *end while*
22.     Output results for post-processing

**Table 1**
Classical benchmark functions (F01-F10).

| Function name | Formula | Bounds |
|---|---|---|
| Sphere | $F01 = \sum_{i=1}^{d} x_i^2$ | $[-100, 100]^d$ |
| Sum of different powers | $F02 = \sum_{i=1}^{d} |x_i|^{i+1}$ | $[-100, 100]^d$ |
| Rotated hyper-ellipsoid | $F03 = \sum_{i=1}^{d} \sum_{j=1}^{i} x_j^2$ | $[-65, 65]^d$ |
| Griewank | $F04 = \sum_{i=1}^{d} \frac{x_i^2}{4000} - \prod_{i=1}^{d} \cos\left(\frac{x_i}{\sqrt{i}}\right) + 1$ | $[-600, 600]^d$ |
| Trid | $F05 = \sum_{i=1}^{d} (x_i - 1)^2 - \sum_{i=2}^{d} x_i x_{i-1}$ | $[-d^2, d^2]^d$ |
| Rastrigin | $F06 = 10d + \sum_{i=1}^{d} \left[x_i^2 - 10\cos(2\pi x_i)\right]$ | $[-5.12, 5.12]^d$ |
| Levy | $F07 = \sin^2(\pi w_1) + \sum_{i=1}^{d-1} (w_i - 1)^2 \left[1 + 10\sin^2(\pi w_i + 1)\right] + (w_d - 1)^2 \left[1 + 10\sin^2(\pi w_d)\right]$ where $w_i = 1 + (x_i - 1)/4$ | $[-5.12, 5.12]^d$ |
| Ackley | $F08 = -20\exp\left(-0.2\sqrt{\frac{1}{d}\sum_{i=1}^{d} x_i^2}\right) - \exp\left(\frac{1}{d}\sum_{i=1}^{d} \cos(2\pi x_i)\right) + 20 + \exp(1)$ | $[-32, 32]^d$ |
| Schwefel | $F09 = 418.9829d - \sum_{i=1}^{d} x_i \sin\left(\sqrt{|x_i|}\right)$ | $[-500, 500]^d$ |
| Rosenbrock | $F10 = \sum_{i=1}^{d-1} \left[100\left(x_{i+1} - x_i^2\right)^2 + \left(x_i - 1\right)^2\right]$ | $[-10, 10]^d$ |

**Table 1 continue**

Classical benchmark functions (F11-F20).

| Function name | Formula | Bounds |
|---|---|---|
| Zakharov | $F11 = \sum_{i=1}^{d} x_i^2 + \left(\sum_{i=1}^{d} 0.5ix_i\right)^2 + \left(\sum_{i=1}^{d} 0.5ix_i\right)^4$ | $[-5, 10]^d$ |
| Dixon-price | $F12 = (x_1 - 1)^2 + \sum_{i=2}^{d} i\left(2x_i^2 - x_{i-1}\right)^2$ | $[-10, 10]^d$ |
| Michalewicz | $F13 = -\sum_{i=1}^{d} \sin(x_i) \sin^{20}\left(\frac{ix_i^2}{\pi}\right)$ | $[0, \pi]^d$ |
| Powell | $F14 = \sum_{i=1}^{d/4}\left[\left(x_{4i-3} + 10x_{4i-2}\right)^2 + 5\left(x_{4i-1} - x_{4i}\right)^2 + \left(x_{4i-2} - 2x_{4i-1}\right)^4 + 10\left(x_{4i-3} + x_{4i}\right)^4\right]$ | $[-10, 10]^d$ |
| Bent cigar | $F15 = x_1^2 + 10^6 \sum_{i=2}^{d} x_i^2$ | $[-10, 10]^d$ |
| Alpine | $F16 = \sum_{i=1}^{d} \left|x_i \sin(x_i) + 0.1x_i\right|$ | $[-10, 10]^d$ |
| Weierstrass | $F17 = \sum_{i=1}^{d}\left(\sum_{k=0}^{20}\left[0.5^k \cos\left(2\pi \cdot 3^k (x_i + 0.5)\right)\right]\right) - d\sum_{k=0}^{20}\left[0.5^k \cos\left(2\pi \cdot 3^k \cdot 0.5\right)\right]$ | $[-0.9, 0.9]^d$ |
| Styblinski-Tang | $F18 = 0.5\sum_{i=1}^{d}\left(x_i^4 - 16x_i^2 + 5x_i\right) + 39.16599d$ | $[-10, 10]^d$ |
| Salomon | $F19 = 1 - \cos\left(2\pi \sum_{i=1}^{d} x_i\right) + 0.1\sum_{i=1}^{d} x_i^2$ | $[-100, 100]^d$ |
| Schaffer F7 | $F20 = \frac{1}{d-1}\sum_{i=1}^{d-1}\left[\left(x_i^2 + x_{i+1}^2\right)^{0.25} + \left(x_i^2 + x_{i+1}^2\right)^{0.25} \sin^2\left(50\left(x_i^2 + x_{i+1}^2\right)^{0.1}\right)\right]$ | $[-100, 100]^d$ |

**Table 2**
Comparison between dBA and classical algorithm on benchmark functions (*F*01~*F*10).

| Function | | dBA | BA | PSO | HS | CS | GA | DE |
|---|---|---|---|---|---|---|---|---|
| F01 | Best | **1.927E-03** | 3.052E-01 | 1.118E+03 | 5.919E+03 | 2.340E+02 | 5.517E+00 | 2.481E+01 |
| | Median | **1.408E-02** | 5.480E+04 | 2.554E+03 | 9.621E+03 | 4.357E+02 | 6.560E+02 | 4.120E+01 |
| | Worst | **2.233E+00** | 6.569E+04 | 5.626E+03 | 1.568E+04 | 6.119E+02 | 7.964E+03 | 8.028E+01 |
| | Mean | **2.256E-01** | 4.920E+04 | 2.852E+03 | 9.618E+03 | 4.153E+02 | 1.678E+03 | 4.411E+01 |
| | SD | 4.869E-01 | 1.859E+04 | 1.105E+03 | 2.226E+03 | 9.518E+01 | 2.032E+03 | 1.259E+01 |
| F02 | Best | 1.011E+06 | 3.313E+09 | 1.609E+20 | 2.573E+33 | 3.229E+17 | **7.488E+04** | 9.080E+08 |
| | Median | **8.171E+09** | 1.294E+45 | 1.085E+28 | 7.580E+37 | 7.654E+19 | 9.245E+29 | 1.177E+11 |
| | Worst | 1.713E+13 | 5.893E+50 | 1.724E+34 | 8.664E+42 | 2.433E+22 | 2.390E+41 | **1.553E+12** |
| | Mean | 1.363E+12 | 4.310E+49 | 1.046E+33 | 3.533E+41 | 2.263E+21 | 1.049E+40 | **3.051E+11** |
| | SD | 4.261E+12 | 1.461E+50 | 3.737E+33 | 1.697E+42 | 5.976E+21 | 4.671E+40 | 4.102E+11 |
| F03 | Best | **1.634E-02** | 8.563E+00 | 4.828E+03 | 4.124E+04 | 1.062E+03 | 8.280E+01 | 9.877E+01 |
| | Median | **3.115E-01** | 2.996E+05 | 1.383E+04 | 5.220E+04 | 1.996E+03 | 5.373E+03 | 1.618E+02 |
| | Worst | **1.256E+02** | 4.370E+05 | 3.416E+04 | 7.472E+04 | 3.409E+03 | 3.294E+04 | 3.850E+02 |
| | Mean | **1.461E+01** | 2.612E+05 | 1.562E+04 | 5.336E+04 | 2.138E+03 | 8.130E+03 | 1.742E+02 |
| | SD | 3.456E+01 | 1.348E+05 | 7.676E+03 | 8.132E+03 | 5.493E+02 | 8.472E+03 | 6.173E+01 |
| F04 | Best | **5.049E-03** | 3.210E+02 | 3.041E+01 | 4.375E+01 | 3.026E+00 | 1.080E-01 | 9.989E-03 |
| | Median | **8.544E-02** | 5.949E+02 | 7.258E+01 | 8.306E+01 | 4.448E+00 | 1.507E+01 | 8.997E-02 |
| | Worst | **5.630E-01** | 6.848E+02 | 1.684E+02 | 1.201E+02 | 6.797E+00 | 5.574E+01 | 2.136E+00 |
| | Mean | **1.405E-01** | 5.816E+02 | 7.481E+01 | 8.040E+01 | 4.567E+00 | 1.900E+01 | 2.303E-01 |
| | SD | 1.481E-01 | 7.884E+01 | 2.717E+01 | 1.588E+01 | 9.934E-01 | 1.828E+01 | 4.210E-01 |
| F05 | Best | 1.685E+03 | 2.967E+06 | 3.078E+05 | 5.169E+05 | 2.831E+04 | 6.326E+03 | **-3.276E+03** |
| | Median | 3.553E+04 | 4.529E+06 | 5.827E+05 | 8.395E+05 | 4.084E+04 | 2.920E+05 | **3.007E+03** |
| | Worst | 9.707E+04 | 5.495E+06 | 1.223E+06 | 1.329E+06 | 8.620E+04 | 7.001E+05 | **2.215E+04** |
| | Mean | 3.423E+04 | 4.436E+06 | 6.204E+05 | 8.815E+05 | 4.242E+04 | 3.194E+05 | **4.901E+03** |
| | SD | 2.590E+04 | 6.360E+05 | 2.312E+05 | 1.932E+05 | 1.118E+04 | 1.916E+05 | 6.627E+03 |
| F06 | Best | 6.812E+01 | 2.420E+02 | 1.707E+02 | 1.330E+02 | 1.129E+02 | **2.994E+01** | 2.998E+01 |
| | Median | 1.057E+02 | 3.074E+02 | 2.517E+02 | 1.625E+02 | 1.378E+02 | **5.895E+01** | 1.575E+02 |
| | Worst | 2.471E+02 | 3.670E+02 | 3.456E+02 | 1.845E+02 | 1.644E+02 | **9.913E+01** | 2.047E+02 |
| | Mean | 1.193E+02 | 3.086E+02 | 2.599E+02 | 1.580E+02 | 1.366E+02 | **5.746E+01** | 1.551E+02 |
| | SD | 4.023E+01 | 3.603E+01 | 3.756E+01 | 1.558E+01 | 1.349E+01 | 1.825E+01 | 3.368E+01 |
| F07 | Best | 1.518E+00 | 3.024E+01 | 2.126E+01 | 1.366E+01 | 2.414E+00 | 1.093E+00 | **1.053E+00** |
| | Median | 4.901E+00 | 6.876E+01 | 3.604E+01 | 2.384E+01 | 4.475E+00 | 4.073E+00 | **1.928E+00** |
| | Worst | 9.997E+00 | 1.135E+02 | 8.057E+01 | 3.540E+01 | 8.813E+00 | 1.562E+01 | **3.388E+00** |
| | Mean | 4.716E+00 | 7.176E+01 | 3.979E+01 | 2.417E+01 | 5.153E+00 | 5.675E+00 | **2.017E+00** |
| | SD | 1.826E+00 | 1.927E+01 | 1.681E+01 | 5.004E+00 | 1.865E+00 | 3.920E+00 | 5.223E-01 |
| F08 | Best | 3.214E+00 | 1.996E+01 | 1.252E+01 | 1.338E+01 | 8.691E+00 | 2.595E+00 | **2.302E+00** |
| | Median | 5.681E+00 | 1.996E+01 | 1.462E+01 | 1.559E+01 | 1.200E+01 | 5.744E+00 | **3.191E+00** |
| | Worst | 8.801E+00 | 1.996E+01 | 1.737E+01 | 1.640E+01 | 1.750E+01 | 1.145E+01 | **3.648E+00** |
| | Mean | 5.839E+00 | 1.996E+01 | 1.474E+01 | 1.540E+01 | 1.209E+01 | 5.920E+00 | **3.191E+00** |
| | SD | 1.730E+00 | 7.062E-04 | 1.235E+00 | 7.839E-01 | 1.753E+00 | 2.453E+00 | 2.904E-01 |
| F09 | Best | 2.895E+03 | 5.685E+03 | 7.293E+03 | **2.281E+03** | 4.522E+03 | 2.736E+03 | 4.745E+03 |
| | Median | 4.492E+03 | 9.365E+03 | 8.803E+03 | **3.698E+03** | 5.045E+03 | 4.228E+03 | 5.370E+03 |
| | Worst | 5.646E+03 | 1.017E+04 | 9.480E+03 | **4.624E+03** | 5.426E+03 | 5.993E+03 | 6.006E+03 |

|  |  |  |  |  |  |  |  |
|---|---|---|---|---|---|---|---|
|  | Mean | 4.357E+03 | 8.940E+03 | 8.712E+03 | **3.722E+03** | 5.056E+03 | 4.208E+03 | 5.407E+03 |
|  | SD | 6.414E+02 | 1.242E+03 | 5.463E+02 | 5.060E+02 | 1.747E+02 | 7.320E+02 | 3.363E+02 |
|  | Best | **2.911E+01** | 3.336E+01 | 8.566E+03 | 8.437E+04 | 6.691E+02 | 1.048E+02 | 4.637E+02 |
|  | Median | **1.038E+02** | 2.473E+02 | 5.394E+04 | 1.588E+05 | 9.105E+02 | 2.756E+03 | 6.892E+02 |
| F10 | Worst | **1.011E+03** | 2.944E+03 | 2.811E+05 | 2.346E+05 | 2.290E+03 | 4.793E+04 | 1.304E+03 |
|  | Mean | **1.645E+02** | 4.916E+02 | 8.159E+04 | 1.597E+05 | 1.073E+03 | 5.961E+03 | 7.193E+02 |
|  | SD | 1.926E+02 | 6.275E+02 | 6.481E+04 | 4.048E+04 | 3.967E+02 | 9.588E+03 | 2.121E+02 |

**Table 3**

Comparison between dBA and classical algorithm on benchmark functions (*F*11~*F*20).

| Function |  | dBA | BA | PSO | HS | CS | GA | DE |
|---|---|---|---|---|---|---|---|---|
|  | Best | 7.536E+01 | **4.799E+00** | 5.754E+02 | 2.960E+02 | 1.337E+02 | 1.122E+01 | 1.414E+02 |
|  | Median | 1.561E+02 | **3.103E+01** | 1.108E+03 | 4.023E+02 | 2.190E+02 | 7.707E+06 | 1.879E+02 |
| F11 | Worst | 2.506E+02 | **1.334E+02** | 1.616E+03 | 5.713E+02 | 3.009E+02 | 9.316E+08 | 2.352E+02 |
|  | Mean | 1.515E+02 | **4.629E+01** | 1.054E+03 | 4.052E+02 | 2.214E+02 | 9.884E+07 | 1.937E+02 |
|  | SD | 4.105E+01 | 3.971E+01 | 2.706E+02 | 6.898E+01 | 4.094E+01 | 2.076E+08 | 2.433E+01 |
|  | Best | **7.448E-01** | 1.323E+00 | 6.914E+03 | 4.057E+04 | 1.059E+02 | 1.207E+01 | 2.650E+01 |
|  | Median | **5.528E+00** | 2.187E+01 | 3.195E+04 | 7.688E+04 | 2.200E+02 | 1.836E+03 | 6.164E+01 |
| F12 | Worst | **1.044E+02** | 9.385E+02 | 1.202E+05 | 1.282E+05 | 6.159E+02 | 3.631E+04 | 1.438E+02 |
|  | Mean | **1.911E+01** | 1.181E+02 | 3.864E+04 | 7.853E+04 | 2.611E+02 | 6.494E+03 | 6.790E+01 |
|  | SD | **2.917E+01** | 2.293E+02 | 2.495E+04 | 2.813E+04 | 1.384E+02 | 9.520E+03 | 2.559E+01 |
|  | Best | -2.094E+01 | -9.637E+00 | -1.307E+01 | -1.477E+01 | -1.673E+01 | **-2.473E+01** | -1.259E+01 |
|  | Median | -1.470E+01 | -8.037E+00 | -9.785E+00 | -1.396E+01 | -1.444E+01 | **-2.154E+01** | -1.123E+01 |
| F13 | Worst | -1.017E+01 | -6.880E+00 | -7.011E+00 | -1.208E+01 | -1.333E+01 | **-1.871E+01** | -1.011E+01 |
|  | Mean | -1.495E+01 | -8.179E+00 | -9.620E+00 | -1.376E+01 | -1.455E+01 | **-2.186E+01** | -1.121E+01 |
|  | SD | 3.135E+00 | 6.847E-01 | 1.732E+00 | 7.115E-01 | 7.923E-01 | **1.793E+00** | 7.085E-01 |
|  | Best | **1.344E+00** | 8.475E+00 | 2.180E+03 | 1.269E+04 | 1.778E+02 | 7.206E+01 | 2.073E+03 |
|  | Median | **2.815E+01** | 4.501E+01 | 7.595E+03 | 3.006E+04 | 3.536E+02 | 1.455E+03 | 4.076E+03 |
| F14 | Worst | **1.918E+02** | 7.523E+02 | 3.698E+04 | 5.492E+04 | 7.081E+02 | 6.457E+03 | 9.651E+03 |
|  | Mean | **4.898E+01** | 1.695E+02 | 9.522E+03 | 3.091E+04 | 3.554E+02 | 2.093E+03 | 4.532E+03 |
|  | SD | 5.028E+01 | 2.220E+02 | 7.144E+03 | 1.079E+04 | 1.311E+02 | 1.975E+03 | 1.861E+03 |
|  | Best | **4.499E+01** | 1.382E+04 | 3.980E+07 | 6.110E+07 | 1.542E+08 | 1.474E+06 | 1.283E+05 |
|  | Median | **3.283E+02** | 4.133E+05 | 7.507E+07 | 8.407E+07 | 3.709E+08 | 1.061E+07 | 2.388E+05 |
| F15 | Worst | **2.518E+03** | 1.247E+07 | 1.557E+08 | 1.111E+08 | 6.179E+08 | 8.335E+07 | 3.528E+05 |
|  | Mean | **4.926E+02** | 1.929E+06 | 8.307E+07 | 8.696E+07 | 3.760E+08 | 1.803E+07 | 2.392E+05 |
|  | SD | 5.304E+02 | 3.115E+06 | 3.129E+07 | 1.489E+07 | 1.192E+08 | 2.213E+07 | 6.522E+04 |
|  | Best | **3.462E-02** | 6.731E+00 | 1.609E+01 | 9.950E+00 | 1.112E+01 | 1.029E-01 | 1.271E+01 |
|  | Median | 3.239E+00 | 1.505E+01 | 2.714E+01 | 1.437E+01 | 1.404E+01 | **7.002E-01** | 1.522E+01 |
| F16 | Worst | 2.046E+01 | 2.832E+01 | 3.970E+01 | 1.846E+01 | 1.834E+01 | **4.277E+00** | 1.912E+01 |
|  | Mean | 3.716E+00 | 1.647E+01 | 2.690E+01 | 1.433E+01 | 1.452E+01 | **1.305E+00** | 1.528E+01 |
|  | SD | 4.409E+00 | 5.854E+00 | 6.382E+00 | 2.381E+00 | 1.746E+00 | 1.157E+00 | 1.682E+00 |
|  | Best | 2.719E+01 | 3.067E+01 | 8.767E+00 | 2.230E+01 | 2.157E+01 | **2.534E+00** | 1.483E+01 |
|  | Median | 3.085E+01 | 3.181E+01 | 3.105E+01 | 2.809E+01 | 2.788E+01 | **6.883E+00** | 2.232E+01 |
| F17 | Worst | 3.320E+01 | 3.270E+01 | 3.283E+01 | 3.144E+01 | 2.988E+01 | **1.464E+01** | 2.805E+01 |
|  | Mean | 3.053E+01 | 3.178E+01 | 2.885E+01 | 2.776E+01 | 2.718E+01 | **7.684E+00** | 2.276E+01 |
|  | SD | 1.668E+00 | 4.720E-01 | 5.410E+00 | 2.174E+00 | 2.463E+00 | 3.387E+00 | 3.374E+00 |
|  | Best | **1.131E+02** | 1.637E+02 | 5.202E+02 | 3.915E+02 | 2.564E+02 | 2.611E+02 | 2.948E+02 |
|  | Median | **1.979E+02** | 2.550E+02 | 6.376E+02 | 7.343E+02 | 3.163E+02 | 3.582E+02 | 3.687E+02 |
| F18 | Worst | **2.686E+02** | 4.179E+02 | 9.585E+02 | 1.137E+03 | 3.596E+02 | 6.594E+02 | 4.154E+02 |
|  | Mean | **1.959E+02** | 2.651E+02 | 6.776E+02 | 7.305E+02 | 3.168E+02 | 3.621E+02 | 3.627E+02 |
|  | SD | 3.767E+01 | 6.899E+01 | 1.252E+02 | 1.644E+02 | 2.533E+01 | 8.306E+01 | 3.236E+01 |
|     | Best   | **3.554E-01** | 5.082E-01 | 4.302E+02 | 6.724E+02 | 2.426E+01 | 9.307E+00 | 3.377E+00 |
|     | Median | **1.328E+00** | 5.697E+03 | 9.703E+02 | 9.243E+02 | 4.450E+01 | 1.757E+02 | 5.082E+00 |
| F19 | Worst  | **2.357E+00** | 7.542E+03 | 2.292E+03 | 1.277E+03 | 8.646E+01 | 7.373E+02 | 8.308E+00 |
|     | Mean   | **1.417E+00** | 5.172E+03 | 1.009E+03 | 9.074E+02 | 4.591E+01 | 2.319E+02 | 5.193E+00 |
|     | SD     | 4.826E-01 | 1.981E+03 | 4.147E+02 | 1.325E+02 | 1.265E+01 | 1.940E+02 | 1.241E+00 |
|     | Best   | 3.861E+00 | 5.453E+00 | 5.735E+00 | 5.296E+00 | 5.729E+00 | **7.870E-01** | 2.769E+00 |
|     | Median | 5.319E+00 | 5.964E+00 | 6.729E+00 | 5.877E+00 | 6.177E+00 | **2.228E+00** | 3.347E+00 |
| F20 | Worst  | 6.766E+00 | 6.693E+00 | 7.411E+00 | 6.838E+00 | 6.674E+00 | **3.680E+00** | 3.761E+00 |
|     | Mean   | 5.262E+00 | 6.019E+00 | 6.617E+00 | 5.946E+00 | 6.187E+00 | **2.064E+00** | 3.324E+00 |
|     | SD     | 7.905E-01 | 3.268E-01 | 5.041E-01 | 3.786E-01 | 2.475E-01 | 9.230E-01 | 2.582E-01 |

## Table 4

Pairwise comparison results (Experiment 1).

| dBA vs | BA | PSO | HS | CS | GA | DE |
|---|---|---|---|---|---|---|
| Wins / Losses | 18/2* | 18/2* | 17/3* | 18/2* | 13/7 | 13/7 |
| Sign test $p$-value | **4.005E-05** | **4.005E-05** | **4.025E-04** | **4.005E-05** | 1.153E-01 | 1.153E-01 |
| Wilcoxon $p$-value | **3.385E-04** | **1.204E-04** | **6.806E-04** | **1.629E-04** | **9.996E-03** | 5.691E-02 |

*Level of significant $\alpha = 0.05$

## Table 5

Friedman, Aligned Friedman and Quade ranks (Experiment 1).

| Algorithm | Friedman | Aligned Friedman | Quade |
|---|---|---|---|
| dBA | 1.85 | 50.90 | 1.58 |
| BA | 5.40 | 93.20 | 4.99 |
| PSO | 5.65 | 89.05 | 5.45 |
| HS | 5.30 | 90.35 | 5.64 |
| CS | 3.65 | 60.45 | 3.69 |
| GA | 3.40 | 56.55 | 4.10 |
| DE | 2.75 | 53.00 | 2.55 |
| Statistic | 55.89 | 16.06 | 11.63 |
| $p$-value | 3.51E-10 | 0.01345 | 3.91E-10 |

## Table 6

Results of post-hoc procedures over all algorithms with dBA as control method at $\alpha=0.05$ (Experiment 1).

| Procedure | $i$ | Algorithm | $z$-value | $p$-value | $p_{Holl}$ | $p_{Rom}$ | $p_{Finn}$ | $p_{Li}$ |
|---|---|---|---|---|---|---|---|---|
|  | 1 | PSO | 5.562630 | 2.66E-08 | 1.59E-07 | 1.52E-07 | 1.59E-07 | 3.27E-08 |
|  | 2 | BA | 5.196668 | 2.03E-07 | 1.01E-06 | 9.65E-07 | 6.09E-07 | 2.50E-07 |
| Friedman | 3 | HS | 5.050283 | 4.41E-07 | 1.76E-06 | 1.68E-06 | 8.82E-07 | 5.43E-07 |
|  | 4 | CS | 2.634930 | 0.008415 | 0.025035 | 0.025246 | 0.012597 | 0.010254 |
|  | 5 | GA | 2.268968 | 0.023270 | 0.045999 | 0.046541 | 0.027859 | 0.027849 |
|  | 6 | DE | 1.317465 | 0.187683 | 0.187683 | 0.187683 | 0.187683 | 0.187683 |
|  | 1 | BA | 3.298051 | 0.000974 | 0.005827 | 0.005554 | 0.005827 | 0.007430 |
| Aligned | 2 | HS | 3.075842 | 0.002099 | 0.010451 | 0.009981 | 0.006284 | 0.015883 |
| Friedman | 3 | PSO | 2.974484 | 0.002935 | 0.011688 | 0.011194 | 0.006284 | 0.022067 |
|  | 4 | CS | 0.744596 | 0.456516 | 0.839469 | 0.869941 | 0.599336 | 0.778274 |

|  | 5 | GA | 0.440520 | 0.659561 | 0.884101 | 0.869941 | 0.725560 | 0.835289 |
|  | 6 | DE | 0.163733 | 0.869941 | 0.884101 | 0.869941 | 0.869941 | 0.869941 |
|  | 1 | HS | 3.005520 | 0.002651 | 0.015803 | 0.015126 | 0.015803 | 0.005014 |
|  | 2 | PSO | 2.867943 | 0.004131 | 0.020487 | 0.019645 | 0.015803 | 0.007792 |
| Quade | 3 | BA | 2.525766 | 0.011545 | 0.045385 | 0.044032 | 0.022956 | 0.021474 |
|  | 4 | GA | 1.866103 | 0.062027 | 0.174777 | 0.186081 | 0.091582 | 0.105470 |
|  | 5 | CS | 1.562729 | 0.118116 | 0.222281 | 0.236233 | 0.140010 | 0.183357 |
|  | 6 | DE | 0.716104 | 0.473927 | 0.473927 | 0.473927 | 0.473927 | 0.473927 |

**Table 7**

Contrast estimation results (experiment 1).

|  | dBA | BA | PSO | HS | CS | GA | DE |
|---|---|---|---|---|---|---|---|
| dBA | 0.000 | **295.9** | **650.2** | **409.4** | **165.8** | **220.9** | **78.14** |
| BA | **-295.9** | 0.000 | 354.3 | 113.5 | -130.1 | -74.99 | -217.8 |
| PSO | **-650.2** | -354.3 | 0.000 | -240.8 | -484.4 | -429.3 | -572.1 |
| HS | **-409.4** | -113.5 | 240.8 | 0.000 | -243.5 | -188.4 | -331.2 |
| CS | **-165.8** | 130.1 | 484.4 | 243.5 | 0.000 | 55.10 | -87.68 |
| GA | **-220.9** | 74.99 | 429.3 | 188.4 | -55.10 | 0.000 | -142.8 |
| DE | **-78.14** | 217.8 | 572.1 | 331.2 | 87.68 | 142.8 | 0.000 |

**Table 8**

Test problems of the CEC'2005 special session on real parameters optimization (Suganthan et al., 2005).

| Functions | Bound | $f_{min}$ |
|---|---|---|
| Unimodal Functions | | |
|     SF01  Shifted Sphere Function | [-100, 100] | -450 |
|     SF02  Shifted Schwefel's Problem 1.2 | [-100, 100] | -450 |
|     SF03  Shifted Rotated High Conditioned Elliptic Function | [-100, 100] | -450 |
|     SF04  Shifted Schwefel's Problem 1.2 with Noise in Fitness | [-100, 100] | -450 |
|     SF05  Schwefel's Problem 2.6 with Global Optimum on Bounds | [-100, 100] | -310 |
| Multimodal Functions | | |
|   Basic Functions | | |
|     SF06  Shifted Rosenbrock's Function | [-100, 100] | 390 |
|     SF07  Shifted Rotated Griewank's Function without Bounds | [0, 600]* | -180 |
|     SF08  Shifted Rotated Ackley's Function with Global Optimum on Bounds | [-32, 32] | -140 |
|     SF09  Shifted Rastrigin's Function | [-5, 5] | -330 |
|     SF10  Shifted Rotated Rastrigin's Function | [-5, 5] | -330 |
|     SF11  Shifted Rotated Weierstrass Function | [-0.5, 0.5] | 90 |
|     SF12  Schwefel's Problem 2.13 | [-$\pi$, $\pi$] | -460 |
|   Expanded Functions | | |
|     SF13  Expanded Extended Griewank's plus Rosenbrock's Function (F8F2) | [-3, 1] | -130 |
|     SF14  Expanded Rotated Extended Scaffe's F6 | [-100, 100] | -300 |
|   Hybrid Composition Functions | | |
|     SF15  Hybrid Composition Function 1 | [-5, 5] | 120 |
|     SF16  Rotated Hybrid Composition Function 1 | [-5, 5] | 120 |
|     SF17  Rotated Hybrid Composition Function 1 with Noise in Fitness | [-5, 5] | 120 |

| | | | |
|---|---|---|---|
| SF18 | Rotated Hybrid Composition Function 2 | [-5, 5] | 10 |
| SF19 | Rotated Hybrid Composition Function 2 with a Narrow Basin for the Global Optimum | [-5, 5] | 10 |
| SF20 | Rotated Hybrid Composition Function 2 with the Global Optimum on the Bounds | [-5, 5] | 10 |
| SF21 | Rotated Hybrid Composition Function 3 | [-5, 5] | 360 |
| SF22 | Rotated Hybrid Composition Function 3 with High Condition Number Matrix | [-5, 5] | 360 |
| SF23 | Non-Continuous Rotated Hybrid Composition Function 3 | [-5, 5] | 360 |
| SF24 | Rotated Hybrid Composition Function 4 | [-5, 5] | 260 |
| SF25 | Rotated Hybrid Composition Function 4 without Bounds | [2, 5]* | 260 |

* Initial range

**Table 9**

Error values achieved for 25 runs of the CEC'2005 benchmark problems.

| Function: | SF01 | SF02 | SF03 | SF04 | SF05 | SF06 | SF07 | SF08 | SF09 |
|---|---|---|---|---|---|---|---|---|---|
| 1st (Best) | 0.000E+00 | 0.000E+00 | 1.426E+04 | 0.000E+00 | 0.000E+00 | 1.254E-02 | 1.232E-02 | 2.025E+01 | 9.950E-01 |
| 7th | 0.000E+00 | 0.000E+00 | 1.037E+05 | 1.023E-12 | 0.000E+00 | 1.572E+00 | 1.872E-01 | 2.035E+01 | 5.970E+00 |
| 13th (Median) | 0.000E+00 | 0.000E+00 | 1.992E+05 | 1.037E-09 | 0.000E+00 | 4.191E+00 | 3.371E-01 | 2.036E+01 | 7.960E+00 |
| 19th | 0.000E+00 | 0.000E+00 | 3.771E+05 | 2.524E-06 | 0.000E+00 | 2.674E+01 | 5.612E-01 | 2.042E+01 | 1.094E+01 |
| 25th (Worst) | 0.000E+00 | 0.000E+00 | 5.250E+05 | 1.649E-03 | 0.000E+00 | 1.139E+03 | 1.139E+00 | 2.049E+01 | 1.393E+01 |
| Mean | 0.000E+00 | 0.000E+00 | 2.389E+05 | 1.134E-04 | 0.000E+00 | 6.646E+01 | 3.870E-01 | 2.038E+01 | 7.880E+00 |
| SD | 0.000E+00 | 0.000E+00 | 1.523E+05 | 3.546E-04 | 0.000E+00 | 2.265E+02 | 2.744E-01 | 5.757E-02 | 3.304E+00 |
| Function: | SF10 | SF11 | SF12 | SF13 | SF14 | SF15 | SF16 | SF17 | SF18 |
| 1st (Best) | 4.379E+00 | 1.934E+00 | 1.725E-02 | 3.447E-01 | 2.015E+00 | 1.140E+02 | 9.683E+01 | 9.886E+01 | 3.000E+02 |
| 7th | 9.108E+00 | 2.706E+00 | 1.001E+01 | 5.705E-01 | 2.769E+00 | 1.654E+02 | 1.086E+02 | 1.124E+02 | 3.480E+02 |
| 13th (Median) | 1.001E+01 | 3.169E+00 | 1.186E+02 | 8.795E-01 | 2.969E+00 | 1.974E+02 | 1.130E+02 | 1.236E+02 | 4.399E+02 |
| 19th | 1.221E+01 | 3.751E+00 | 2.125E+02 | 1.107E+00 | 3.258E+00 | 2.302E+02 | 1.189E+02 | 1.302E+02 | 5.303E+02 |
| 25th (Worst) | 1.731E+01 | 4.613E+00 | 6.420E+02 | 1.699E+00 | 3.470E+00 | 4.108E+02 | 1.366E+02 | 1.474E+02 | 8.006E+02 |
| Mean | 1.087E+01 | 3.223E+00 | 1.545E+02 | 8.937E-01 | 2.950E+00 | 2.075E+02 | 1.151E+02 | 1.219E+02 | 4.853E+02 |
| SD | 3.198E+00 | 7.465E-01 | 1.740E+02 | 3.802E-01 | 3.692E-01 | 6.305E+01 | 1.083E+01 | 1.105E+01 | 1.689E+02 |
| Function: | SF19 | SF20 | SF21 | SF22 | SF23 | SF24 | SF25 | | |
| 1st (Best) | 3.000E+02 | 3.000E+02 | 3.000E+02 | 5.280E+02 | 4.252E+02 | 2.000E+02 | 2.002E+02 | | |
| 7th | 3.000E+02 | 3.000E+02 | 3.183E+02 | 5.322E+02 | 5.539E+02 | 2.000E+02 | 3.741E+02 | | |
| 13th (Median) | 3.693E+02 | 3.560E+02 | 4.189E+02 | 5.381E+02 | 5.540E+02 | 2.000E+02 | 3.786E+02 | | |
| 19th | 4.088E+02 | 3.966E+02 | 5.000E+02 | 7.712E+02 | 5.595E+02 | 2.000E+02 | 3.842E+02 | | |
| 25th (Worst) | 1.039E+03 | 1.038E+03 | 5.004E+02 | 8.011E+02 | 5.836E+02 | 2.000E+02 | 5.000E+02 | | |
| Mean | 4.205E+02 | 3.964E+02 | 4.080E+02 | 6.307E+02 | 5.464E+02 | 2.000E+02 | 3.608E+02 | | |
| SD | 1.847E+02 | 1.677E+02 | 8.121E+01 | 1.203E+02 | 3.703E+01 | 8.343E-13 | 7.884E+01 | | |

**Table10**
Comparison of the average error obtained with dBA and the other algorithms (experiment 2).

| Function | dBA | PSO | IPOP-CMA-ES | CHC | SSGA | SS-BLX | SS-Arit | DE-Bin | DE-Exp | SaDE |
|---|---|---|---|---|---|---|---|---|---|---|
| SF01 | **0.000E+00** | 1.234E-04 | **0.000E+00** | 2.464E+00 | 8.420E-09 | 3.402E+01 | 1.064E+00 | 7.716E-09 | 8.260E-09 | 8.416E-09 |
| SF02 | **0.000E+00** | 2.595E-02 | **0.000E+00** | 1.180E+02 | 8.719E-05 | 1.730E+00 | 5.282E+00 | 8.342E-09 | 8.181E-09 | 8.208E-09 |
| SF03 | 2.356E+05 | 5.174E+04 | **0.000E+00** | 2.699E+05 | 7.948E+04 | 1.844E+05 | 2.535E+05 | 4.233E+01 | 9.935E+01 | 6.560E+03 |
| SF04 | 1.215E-03 | 2.488E+00 | 2.932E+03 | 9.190E+01 | 2.585E-03 | 6.228E+00 | 5.755E+00 | **7.686E-09** | 8.350E-09 | 8.087E-09 |
| SF05 | **0.000E+00** | 4.095E+02 | 8.104E-10 | 2.641E+02 | 1.343E+02 | 2.185E+00 | 1.443E+01 | 8.608E-09 | 8.514E-09 | 8.640E-09 |
| SF06 | 3.538E+01 | 7.310E+02 | **0.000E+00** | 1.416E+06 | 6.171E+00 | 1.145E+02 | 4.945E+02 | 7.956E-09 | 8.391E-09 | 1.612E-02 |
| SF07 | **4.314E-01** | 2.678E+01 | 1.267E+03 | 1.269E+03 | 1.271E+03 | 1.966E+03 | 1.908E+03 | 1.266E+03 | 1.265E+03 | 1.263E+03 |
| SF08 | 2.035E+01 | 2.043E+01 | **2.001E+01** | 2.034E+01 | 2.037E+01 | 2.035E+01 | 2.036E+01 | 2.033E+01 | 2.038E+01 | 2.032E+01 |
| SF09 | 8.216E+00 | 1.438E+01 | 2.841E+01 | 5.886E+00 | **7.286E-09** | 4.195E+00 | 5.960E+00 | 4.549E+00 | 8.151E-09 | 8.330E-09 |
| SF10 | 1.049E+01 | 1.404E+01 | 2.327E+01 | **7.123E+00** | 1.712E+01 | 1.239E+01 | 2.179E+01 | 1.228E+01 | 1.118E+01 | 1.548E+01 |
| SF11 | 3.758E+00 | 5.590E+00 | **1.343E+00** | 1.599E+00 | 3.255E+00 | 2.929E+00 | 2.858E+00 | 2.434E+00 | 2.067E+00 | 6.796E+00 |
| SF12 | 1.885E+02 | 6.362E+02 | 2.127E+02 | 7.062E+02 | 2.794E+02 | 1.506E+02 | 2.411E+02 | 1.061E+02 | 6.309E+01 | **5.634E+01** |
| SF13 | **1.045E+00** | 1.503E+00 | 1.134E+00 | 8.297E+01 | 6.713E+01 | 3.245E+01 | 5.479E+01 | 1.573E+00 | 6.403E+01 | 7.070E+01 |
| SF14 | 2.965E+00 | 3.304E+00 | 3.775E+00 | **2.073E+00** | 2.264E+00 | 2.796E+00 | 2.970E+00 | 3.073E+00 | 3.158E+00 | 3.415E+00 |
| SF15 | 2.166E+02 | 3.398E+02 | 1.934E+02 | 2.751E+02 | 2.920E+02 | 1.136E+02 | 1.288E+02 | 3.722E+02 | 2.940E+02 | **8.423E+01** |
| SF16 | 1.182E+02 | 1.333E+02 | 1.170E+02 | **9.729E+01** | 1.053E+02 | 1.041E+02 | 1.134E+02 | 1.117E+02 | 1.125E+02 | 1.227E+02 |
| SF17 | 1.266E+02 | 1.497E+02 | 3.389E+02 | **1.045E+02** | 1.185E+02 | 1.183E+02 | 1.279E+02 | 1.421E+02 | 1.312E+02 | 1.387E+02 |
| SF18 | **4.471E+02** | 8.512E+02 | 5.570E+02 | 8.799E+02 | 8.063E+02 | 7.668E+02 | 6.578E+02 | 5.097E+02 | 4.482E+02 | 5.320E+02 |
| SF19 | 4.499E+02 | 8.497E+02 | 5.292E+02 | 8.798E+02 | 8.899E+02 | 7.555E+02 | 7.010E+02 | 5.012E+02 | **4.341E+02** | 5.195E+02 |
| SF20 | **3.946E+02** | 8.509E+02 | 5.264E+02 | 8.960E+02 | 8.893E+02 | 7.463E+02 | 6.411E+02 | 4.928E+02 | 4.188E+02 | 4.767E+02 |
| SF21 | **4.135E+02** | 9.138E+02 | 4.420E+02 | 8.158E+02 | 8.522E+02 | 4.851E+02 | 5.005E+02 | 5.240E+02 | 5.420E+02 | 5.140E+02 |
| SF22 | **5.889E+02** | 8.071E+02 | 7.647E+02 | 7.742E+02 | 7.519E+02 | 6.828E+02 | 6.941E+02 | 7.715E+02 | 7.720E+02 | 7.655E+02 |
| SF23 | **5.595E+02** | 1.028E+03 | 8.539E+02 | 1.075E+03 | 1.004E+03 | 5.740E+02 | 5.828E+02 | 6.337E+02 | 5.824E+02 | 6.509E+02 |
| SF24 | **2.000E+02** | 4.120E+02 | 6.101E+02 | 2.959E+02 | 2.360E+02 | 2.513E+02 | 2.011E+02 | 2.060E+02 | 2.020E+02 | **2.000E+02** |
| SF25 | **3.184E+02** | 5.099E+02 | 1.818E+03 | 1.764E+03 | 1.747E+03 | 1.794E+03 | 1.804E+03 | 1.744E+03 | 1.742E+03 | 1.738E+03 |

**Table 11**

Pairwise comparisons results (experiment 2).

| dBA vs. | PSO | IPOP-CMA-ES | CHC | SSGA | SS-BLX | SS-Arit | DE-Bin | DE-Exp | SaDE |
|---|---|---|---|---|---|---|---|---|---|
| Wins | 24 | 17 | 18 | 18 | 16 | 21 | 17 | 17 | 17 |
| Ties | 0 | 2 | 0 | 0 | 1 | 0 | 0 | 0 | 1 |
| Loses | 1 | 6 | 7 | 7 | 8 | 4 | 8 | 8 | 7 |
| $p$-value | **1.55E-06** | **0.03469** | **0.04329** | **0.04329** | 0.15159 | **0.00091** | 0.10775 | 0.10775 | 0.06391 |
| α* | **0.05** | **0.05** | **0.05** | **0.05** | -- | **0.05** | 0.1 | 0.1 | 0.1 |
| Wilcoxon $p$-value | **0.00022** | **0.01497** | **0.00040** | **0.01725** | **0.04867** | **0.00060** | 0.07356 | 0.19190 | 0.10960 |

*Level of significance according to number of wins (Derrac et al., 2011)

**Table 12**

The Friedman, Aligned Friedman and Quade ranks (experiment 2).

| Algorithm | Friedman | Aligned Friedman | Quade |
|---|---|---|---|
| dBA | 3.32 | 85.20 | 3.07 |
| PSO | 7.72 | 150.08 | 7.33 |
| IPOP-CMA-ES | 5.32 | 129.64 | 5.32 |
| CHC | 7.08 | 179.48 | 8.24 |
| SSGA | 6.36 | 150.56 | 6.72 |
| SS-BLX | 5.50 | 124.30 | 5.92 |
| SS-Arit | 6.32 | 127.56 | 6.52 |
| DE-Bin | 4.44 | 102.52 | 4.15 |
| DE-Exp | 4.24 | 102.32 | 3.75 |
| SaDE | 4.70 | 103.34 | 3.99 |
| Statistic | 46.29 | 21.52 | 8.87 |
| $p$-value | 5.32E-07 | 0.01052 | 2.40E-11 |

**Table 13**

Results of post-hoc procedures over all algorithms with dBA as control method at $\alpha=0.05$ (Experiment 2).

| Procedure | $i$ | Algorithm | $z$-value | $p$-value | $p_{Holl}$ | $p_{Rom}$ | $p_{Finn}$ | $p_{Li}$ |
|---|---|---|---|---|---|---|---|---|
| Friedman | 1 | PSO | 5.138093 | 2.78E-07 | 2.50E-06 | 2.37E-06 | 2.50E-06 | 3.87E-07 |
| | 2 | CHC | 4.390734 | 1.13E-05 | 9.04E-05 | 8.59E-05 | 5.08E-05 | 1.57E-05 |
| | 3 | SSGA | 3.549955 | 0.000385 | 0.002694 | 0.002564 | 0.001155 | 0.000537 |
| | 4 | SS-Arit | 3.503245 | 0.000460 | 0.002755 | 0.002622 | 0.001155 | 0.000640 |
| | 5 | SS-BLX | 2.545692 | 0.010906 | 0.053354 | 0.051858 | 0.019545 | 0.014976 |
| | 6 | IPOP-CMA-ES | 2.335497 | 0.019517 | 0.075814 | 0.074441 | 0.029133 | 0.026488 |
| | 7 | SaDE | 1.611493 | 0.107072 | 0.288051 | 0.282675 | 0.135502 | 0.129880 |
| | 8 | DE-Bin | 1.307878 | 0.190915 | 0.345381 | 0.282675 | 0.212059 | 0.210203 |
| | 9 | DE-Exp | 1.074329 | 0.282675 | 0.345381 | 0.282675 | 0.282675 | 0.282675 |
| Aligned Friedman | 1 | CHC | 4.609548 | 4.04E-06 | 3.63E-05 | 3.45E-05 | 3.63E-05 | 6.75E-06 |
| | 2 | SSGA | 3.195588 | 0.001395 | 0.011109 | 0.010071 | 0.006264 | 0.002330 |
| | 3 | PSO | 3.172120 | 0.001513 | 0.011109 | 0.010071 | 0.006264 | 0.002527 |
| | 4 | IPOP-CMA-ES | 2.172765 | 0.029798 | 0.165987 | 0.169999 | 0.065800 | 0.047508 |
| | 5 | SS-Arit | 2.071070 | 0.038352 | 0.177606 | 0.182363 | 0.067972 | 0.060323 |
| | 6 | SS-BLX | 1.911681 | 0.055917 | 0.205598 | 0.213271 | 0.082692 | 0.085586 |
| | 7 | SaDE | 0.886903 | 0.375131 | 0.756013 | 0.402574 | 0.453685 | 0.385716 |
| | 8 | DE-Bin | 0.846811 | 0.397100 | 0.756013 | 0.402574 | 0.453685 | 0.399286 |
| | 9 | DE-Exp | 0.837033 | 0.402574 | 0.756013 | 0.402574 | 0.453685 | 0.402574 |
| Quade | 1 | CHC | 2.486896 | 0.012886 | 0.110175 | 0.110254 | 0.110175 | 0.047820 |
| | 2 | PSO | 2.048467 | 0.040514 | 0.281696 | 0.308134 | 0.169818 | 0.136364 |
| | 3 | SSGA | 1.755194 | 0.079226 | 0.438860 | 0.527272 | 0.219345 | 0.235921 |
| | 4 | SS-Arit | 1.658918 | 0.097132 | 0.458318 | 0.554145 | 0.219345 | 0.274601 |
| | 5 | SS-BLX | 1.372310 | 0.169967 | 0.606017 | 0.743411 | 0.284892 | 0.398463 |
| | 6 | IPOP-CMA-ES | 1.082000 | 0.279253 | 0.730144 | 0.743411 | 0.388108 | 0.521147 |
| | 7 | DE-Bin | 0.519893 | 0.603138 | 0.937495 | 0.743411 | 0.695235 | 0.701546 |
| | 8 | SaDE | 0.442872 | 0.657858 | 0.937495 | 0.743411 | 0.700786 | 0.719405 |
| | 9 | DE-Exp | 0.327340 | 0.743411 | 0.937495 | 0.743411 | 0.743411 | 0.743411 |

**Table 14**

Contrast estimation results (experiment 2).

| | dBA | PSO | IPOP-CMA-ES | CHC | SSGA | SS-BLX | SS-Arit | DE-Bin | DE-Exp | SaDE |
|---|---|---|---|---|---|---|---|---|---|---|
| dBA | 0.000 | **34.97** | **23.51** | **72.87** | **27.00** | **11.61** | **15.95** | **12.13** | **16.75** | **10.16** |
| PSO | **-34.97** | 0.000 | -11.46 | 37.90 | -7.967 | -23.36 | -19.02 | -22.84 | -18.22 | -24.81 |

| | | | | | | | | | |
|---|---|---|---|---|---|---|---|---|---|
| IPOP-CMA-ES | **-23.51** | 11.46 | 0.000 | 49.36 | 3.493 | -11.90 | -7.557 | -11.38 | -6.757 | -13.35 |
| CHC | **-72.87** | -37.90 | -49.36 | 0.000 | -45.87 | -61.26 | -56.92 | -60.74 | -56.12 | -62.71 |
| SSGA | **-27.00** | 7.967 | -3.493 | 45.87 | 0.000 | -15.39 | -11.05 | -14.87 | -10.25 | -16.84 |
| SS-BLX | **-11.61** | 23.36 | 11.90 | 61.26 | 15.39 | 0.000 | 4.342 | 0.523 | 5.143 | -1.448 |
| SS-Arit | **-15.95** | 19.02 | 7.557 | 56.92 | 11.05 | -4.342 | 0.000 | -3.820 | 0.800 | -5.790 |
| DE-Bin | **-12.13** | 22.84 | 11.38 | 60.74 | 14.87 | -0.523 | 3.820 | 0.000 | 4.620 | -1.971 |
| DE-Exp | **-16.75** | 18.22 | 6.757 | 56.12 | 10.25 | -5.143 | -0.800 | -4.620 | 0.000 | -6.591 |
| SaDE | **-10.16** | 24.81 | 13.35 | 62.71 | 16.84 | 1.448 | 5.790 | 1.971 | 6.591 | 0.000 |

**Table 15**

Comparisons between dBA and enhanced versions of BA (experiment 3).

| F | Bounds | $D$ | $N$ | $t_{max}$ | Trials | Present study (dBA) | | Literature | | |
|---|---|---|---|---|---|---|---|---|---|---|
| | | | | | | Mean | SD | Mean | SD | Ref. |
| F01 | [-5.12, 5.12] | 30 | 10000 | 100 | 30 | **1.53E-12** | 1.17E-12 | 7.92E-06 | 8.06E-07 | CBSO (Jordehi, 2015) |
| | [100, 100] | 5 | 30 | 500 | 30 | **5.55E-38** | 2.81E-37 | 1.8518 | 2.4981 | BBA (Mirjalili et al., 2013) |
| | [100, 100] | 30 | 40 | 2000 | 50 | **7.32E-26** | 8.46E-26 | 4.61E-05 | N.A. | SAGBA (He et al., 2014) |
| | [-600, 600] | 10 | 100 | 100 | 25 | **6.61E-06** | 1.17E-05 | 6.98E-01 | 2.62E-01 | HSABA (Fister, Fong, et al., 2014) |
| | [-5.12, 5.12] | 10 | 50 | 2000 | 30 | **0.00E+00** | 0.00E+00 | 1.31E-31 | 3.91E-31 | EnBA (Yilmaz & Küçüksille, 2015) |
| | [-5.12, 5.12] | 60 | 50 | 6000 | 30 | **8.99E-17** | 2.31E-16 | 1.08E+01 | 3.70E+00 | MBA (Yilmaz et al., 2014) |
| F04 | [-600, 600] | 30 | 10000 | 100 | 30 | 1.97E-02 | 1.37E-02 | **0.0104** | 0.0021 | CBSO (Jordehi, 2015) |
| | [-600, 600] | 5 | 30 | 500 | 30 | **9.46E-02** | 7.43E-02 | 0.2463 | 0.0839 | BBA (Mirjalili et al., 2013) |
| | [-600, 600] | 2 | 40 | 2000 | 50 | **0.00E+00** | 0.00E+00 | 3.64E-08 | N.A. | SAGBA (He et al., 2014) |
| | [-600, 600] | 10 | 100 | 100 | 25 | **1.08E-01** | 5.64E-02 | 4.05E-01 | 2.27E-01 | HSABA (Fister, Fong, et al., 2014) |
| | [-600, 600] | 10 | 50 | 2000 | 30 | **1.08E-01** | 6.83E-02 | 9.01E-01 | 6.57E-01 | EnBA (Yilmaz & Küçüksille, 2015) |
| | [-600, 600] | 60 | 50 | 6000 | 30 | **9.81E-02** | 1.04E-01 | 3.19E+02 | 5.64E+01 | MBA (Yilmaz et al., 2014) |
| F06 | [-5.12, 5.12] | 5 | 30 | 500 | 30 | 1.99E+00 | 1.26E+00 | **1.585** | 1.3352 | BBA (Mirjalili et al., 2013) |
| | [-5.12, 5.12]* | 2 | 40 | 2000 | 50 | **0.00E+00** | 0.00E+00 | 1.58E-07 | N.A. | SAGBA (He et al., 2014) |
| | [-15, 15] | 10 | 100 | 100 | 25 | **2.50E+01** | 1.05E+01 | 8.99E+01 | 5.87E+01 | HSABA (Fister, Fong, et al., 2014) |
| | [-5.12, 5.12] | 10 | 50 | 2000 | 30 | **1.00E+01** | 3.43E+00 | 1.01E+01 | 4.14E-00 | EnBA (Yilmaz & Küçüksille, 2015) |
| | [-5.12, 5.12] | 60 | 50 | 6000 | 30 | **3.71E+02** | 3.91E+01 | 3.84E+02 | 1.21E+02 | MBA (Yilmaz et al., 2014) |
| F08 | [-32, 32] | 5 | 30 | 500 | 30 | **1.66E-05** | 4.90E-05 | 1.16E+00 | 7.28E-01 | BBA (Mirjalili et al., 2013) |
| | [-32, 32]* | 2 | 40 | 2000 | 50 | **8.88E-16** | 0.00E+00 | 6.19E-04 | N.A. | SAGBA (He et al., 2014) |
| | [-32, 32] | 10 | 100 | 100 | 25 | **5.79E-01** | 8.71E-01 | 1.84E+01 | 2.00E+01 | HSABA (Fister, Fong, et al., 2014) |
| | [-32, 32]* | 10 | 50 | 2000 | 30 | 4.22E-08 | 4.22E-08 | **4.21E-09** | 2.30E-08 | EnBA (Yilmaz & Küçüksille, 2015) |
| | [-32, 32] | 60 | 50 | 6000 | 30 | **1.09E+01** | 5.28E+00 | 1.45E+01 | 8.07E-01 | MBA (Yilmaz et al., 2014) |
| F10 | [-5, 10] | 30 | 10000 | 100 | 30 | 5.76E+01 | 8.45E+01 | **0.2194** | 0.018 | CBSO (Jordehi, 2015) |
| | [-30, 30] | 5 | 30 | 500 | 30 | **1.54E+00** | 1.76E+00 | 25.0743 | 28.443 | BBA (Mirjalili et al., 2013) |
| | [-30, 30]* | 2 | 40 | 2000 | 50 | **3.94E-32** | 7.89E-32 | 4.68E-06 | N.A. | SAGBA (He et al., 2014) |
| | [-15, 15] | 10 | 100 | 100 | 25 | **2.71E+01** | 4.12E+01 | 1.75E+04 | 7.10E+01 | HSABA (Fister, Fong, et al., 2014) |
| | [30, 30]* | 10 | 50 | 2000 | 30 | 2.59E+00 | 1.97E+00 | **1.32E-01** | 7.27E-01 | EnBA (Yilmaz & Küçüksille, 2015) |
| | [-2,408,2,408] | 60 | 50 | 6000 | 30 | **1.16E+02** | 4.89E+01 | 2.57E+02 | 6.19E+01 | MBA (Yilmaz et al., 2014) |
| F11 | [-5,10] | 30 | 10000 | 100 | 30 | 3.25E+01 | 6.95E+00 | **1.07E-05** | 4.52E-07 | CBSO (Jordehi, 2015) |
| | [-5,10]* | 2 | 40 | 2000 | 50 | **0.00E+00** | 0.00E+00 | 6.12E-10 | N.A | SAGBA (He et al., 2014) |
| | [-5,10] | 10 | 100 | 100 | 25 | **5.33E-01** | 4.93E-01 | 1.45E+01 | 2.20E+00 | HSABA (Fister, Fong, et al., 2014) |
| | [-5,10] | 10 | 50 | 2000 | 30 | **7.24E-43** | 1.15E-42 | 3.49E-11 | 1.91E-10 | EnBA (Yilmaz & Küçüksille, 2015) |

*: data not mentioned in the reference.

**Table 16**
Comparison between dBA and CBSO (experiment 3).

| N | $t_{max}$ | Spherical | | | Zakharov | | |
|---|---|---|---|---|---|---|---|
| | | Mean | SD | Success rate | Mean | SD | Success rate |
| 50 | 20000 | 0.00E+00 | 0.00E+00 | 100.00% | 1.13E-22 | 6.11E-22 | 100.00% |
| 100 | 10000 | 4.25E-273 | 0.00E+00 | 100.00% | 7.39E-24 | 3.97E-23 | 100.00% |
| 500 | 2000 | 6.89E-107 | 1.83E-106 | 100.00% | 1.79E-10 | 2.66E-10 | 56.67% |
| 1000 | 1000 | 2.60E-65 | 3.76E-65 | 100.00% | 9.13E-05 | 9.56E-05 | 0.00% |
| 5000 | 200 | 3.72E-18 | 2.11E-18 | 100.00% | 1.16E+01 | 3.92E+00 | 0.00% |
| 10000 | 100 | 1.42E-12 | 1.25E-12 | 100.00% | 3.20E+01 | 7.68E+00 | 0.00% |
| CBSO(Jordehi, 2015) | | 7.92E-06 | 8.06E-07 | 0.00% | 1.07E-05 | 4.52E-07 | 0.00% |

| N | $t_{max}$ | Griewank | | | Rosenbrock | | |
|---|---|---|---|---|---|---|---|
| | | Mean | SD | Success rate | Mean | SD | Success rate |
| 50 | 20000 | 5.12E-02 | 6.76E-02 | 33.33% | 3.02E+01 | 3.93E+01 | 0.00% |
| 100 | 10000 | 5.07E-02 | 1.15E-01 | 30.00% | 2.49E+01 | 2.43E+01 | 0.00% |
| 500 | 2000 | 1.15E-02 | 1.23E-02 | 33.33% | 3.48E+01 | 3.75E+01 | 0.00% |
| 1000 | 1000 | 1.21E-02 | 1.29E-02 | 30.00% | 4.75E+01 | 7.68E+01 | 0.00% |
| 5000 | 200 | 1.05E-02 | 1.08E-02 | 36.67% | 5.88E+01 | 7.96E+01 | 0.00% |
| 10000 | 100 | 1.63E-02 | 1.80E-02 | 0.00% | 5.96E+01 | 7.45E+01 | 0.00% |
| CBSO(Jordehi, 2015) | | 1.04E-02 | 2.10E-03 | 0.00% | 2.19E-01 | 1.80E-02 | 0.00% |

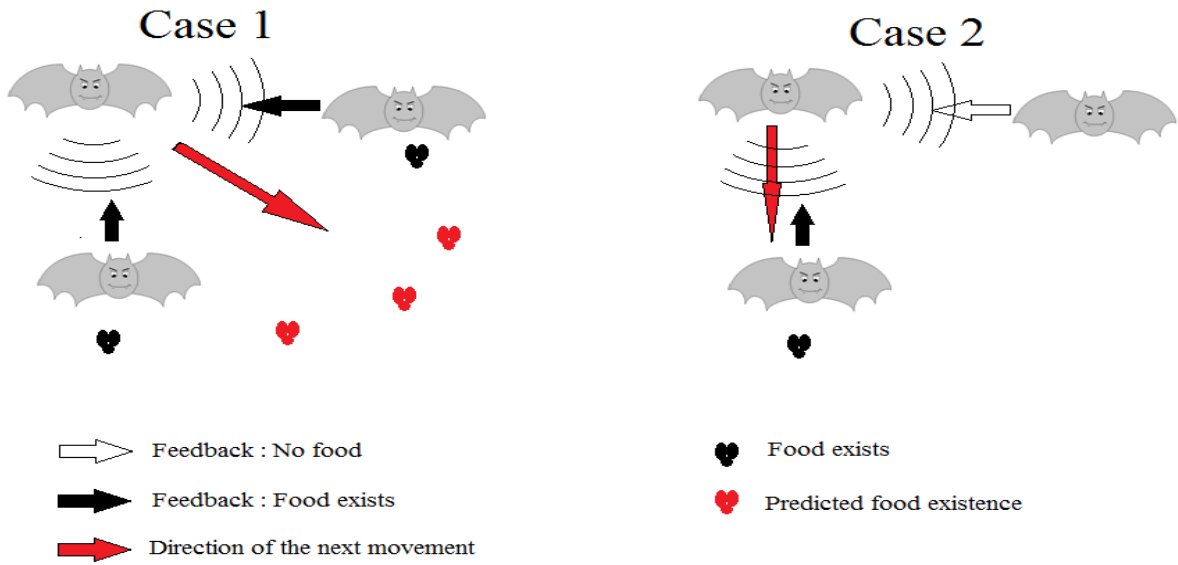

**Fig. 1.** Hypothetical scheme of the directional echolocation behavior.

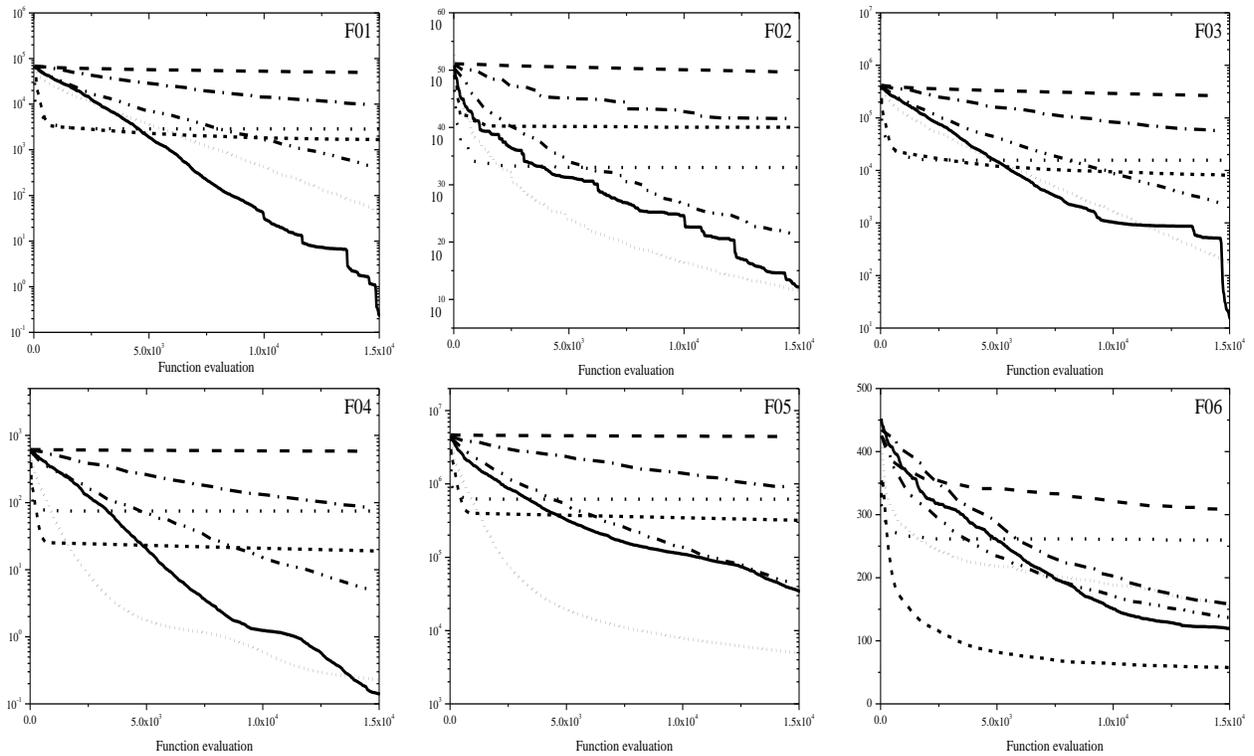

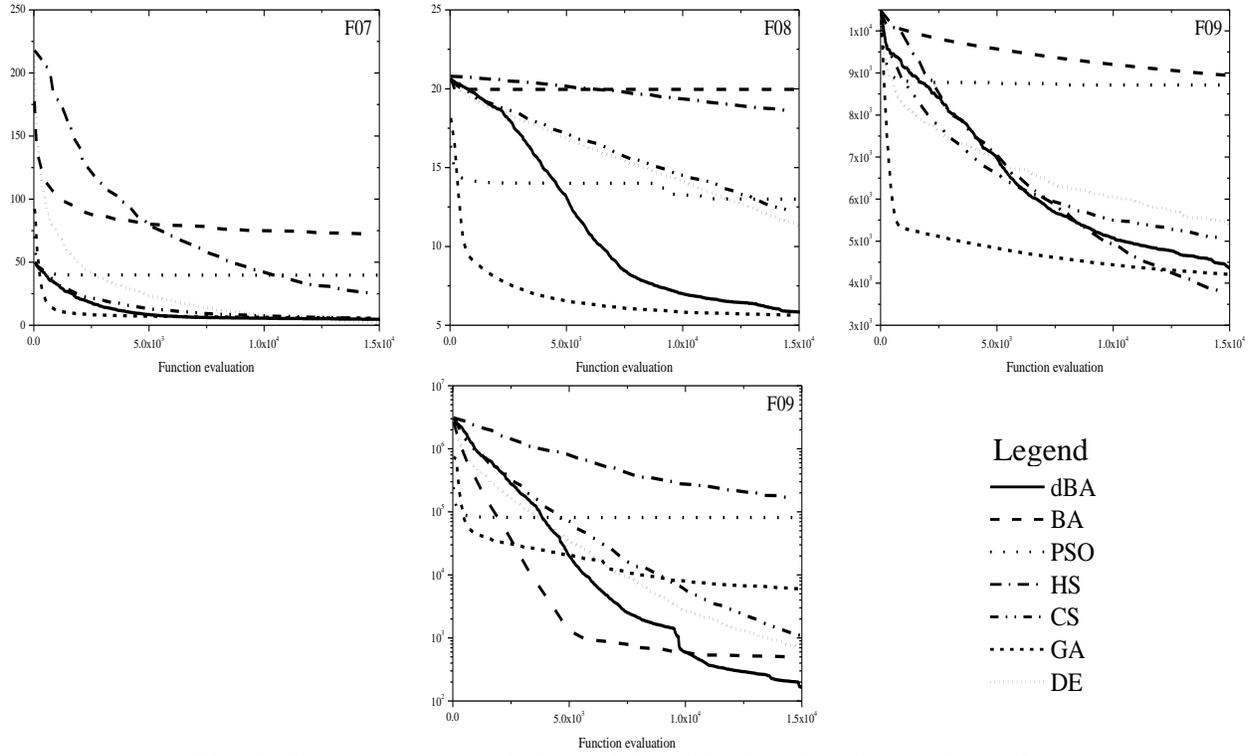

**Fig. 2.** Convergence evolution of the objective function (*F*01-*F*10).

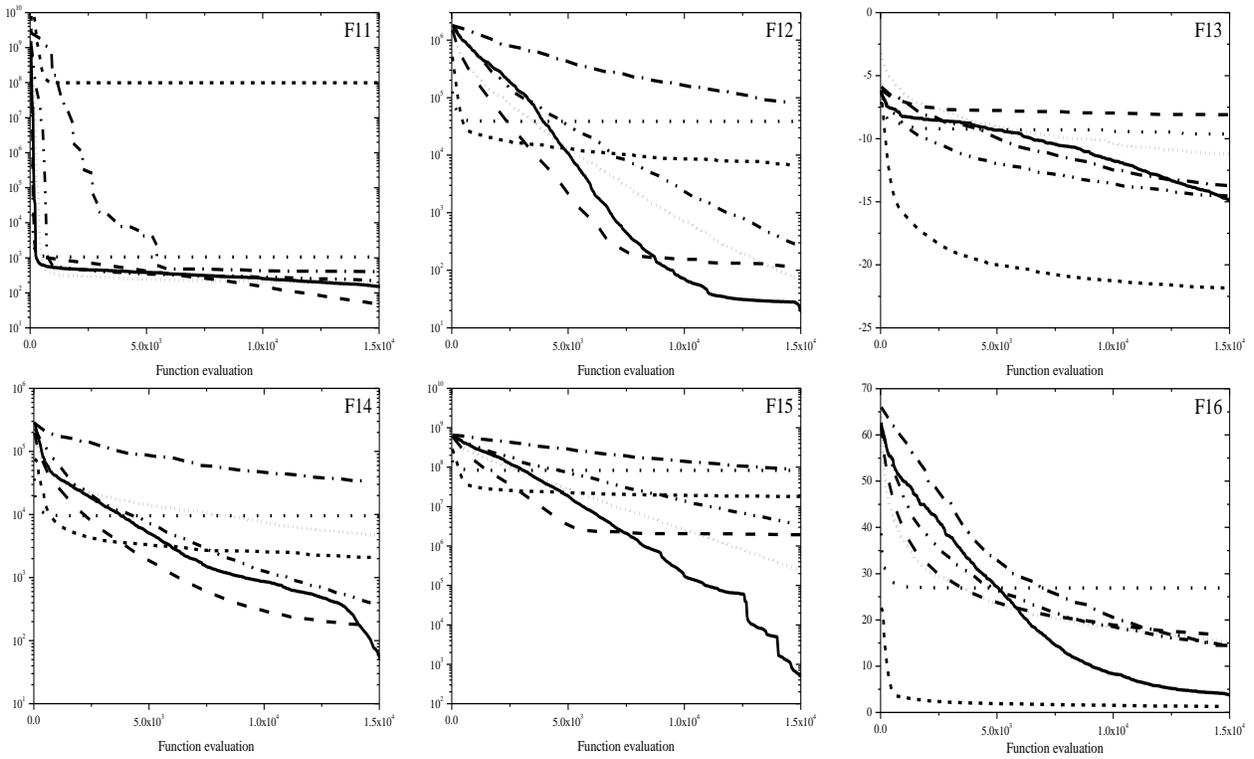

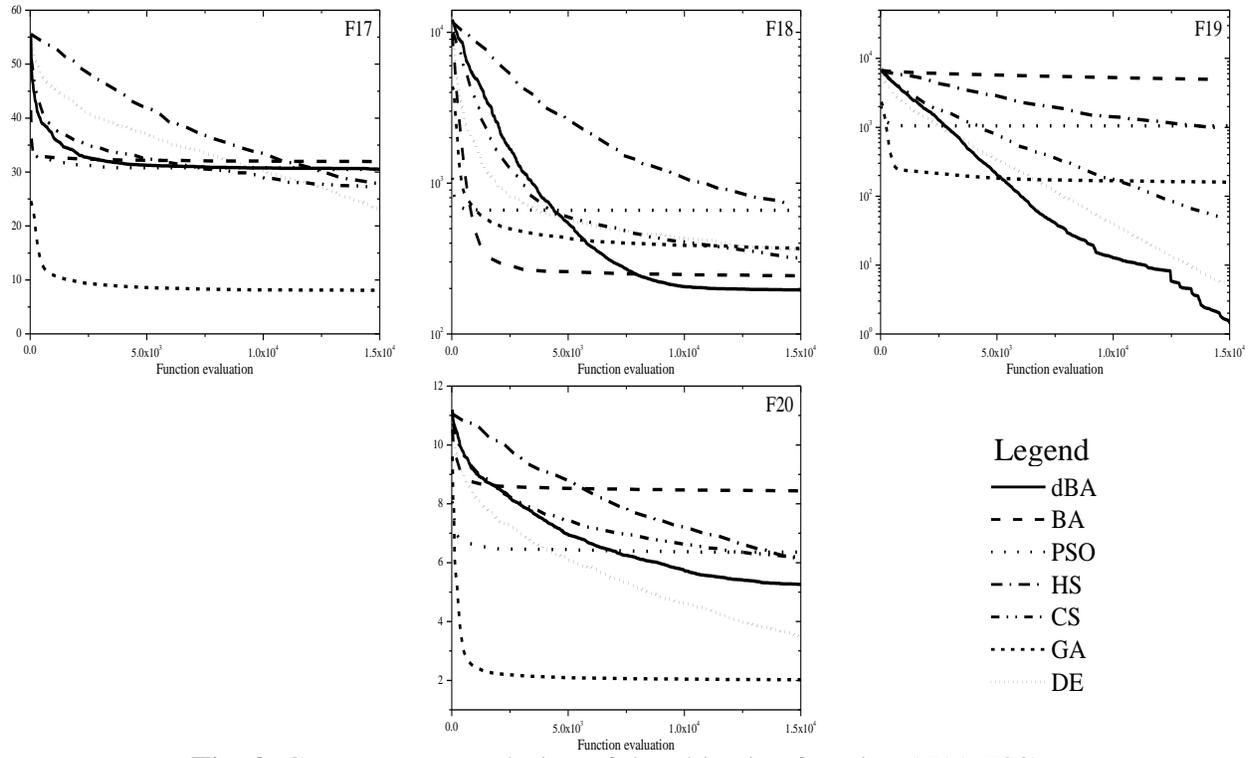

**Fig. 3.** Convergence evolution of the objective function (*F*11-*F*20).